\documentclass{bmvc2k}

\usepackage[utf8]{inputenc} %
\usepackage[T1]{fontenc}    %
\usepackage{hyperref}       %
\usepackage{url}            %
\usepackage{booktabs}       %
\usepackage{amsfonts}       %
\usepackage{nicefrac}       %
\usepackage{microtype}      %
\usepackage{xcolor}         %
\usepackage{graphicx}
\usepackage{xspace} 
\usepackage{amsmath,graphicx}
\usepackage{pifont}
\usepackage{breqn}
\usepackage{amssymb}
\usepackage{xcolor}
\usepackage{nicefrac}
\usepackage{bm}
\usepackage{multirow}
\usepackage{makecell}
\usepackage{tabularx} 
\usepackage{xcolor}
\newcommand{\supp}{the Appendix\xspace}

\def\eg{\emph{e.g.}} 
\def\ie{\emph{i.e.}}

\def\etal{\emph{et al.}}

\def\newpara{\vspace{2pt}}
\def\psec{\vspace{-10pt}}
\def\psubsec{\vspace{-5pt}}
\def\pfig{\vspace{-15pt}}
\def\ptab{\vspace{-7pt}}

\newcommand{\maploc}{${\text{mAP}}^{Loc}$}

\newcommand{\mapcls}{${\text{mAP}}^{Cls}$}
\newcommand{\aconecls}{Acc$^{Cls}_{@1}$}
\newcommand{\ackcls}{Acc$^{Cls}_{@k}$}
\newcommand{\acfivecls}{Acc$^{Cls}_{@5}$}
\newcommand{\losscls}{$L^{Cls}$}

\newcommand{\losslocbce}{$L^{Loc}$}
\newcommand{\losslocsoftmax}{$L^{Loc}_{softmax}$}
\newcommand{\rebuttal}[1]{\textcolor{black}{#1}}

\title{Visual Keyword Spotting with Attention}

\newcommand\blfootnote[1]{%
  \begingroup
  \renewcommand\thefootnote{}\footnote{#1}%
  \addtocounter{footnote}{-1}%
  \endgroup
}

\addauthor{K R Prajwal*}{prajwal@robots.ox.ac.uk}{1}
\addauthor{Liliane Momeni*}{liliane@robots.ox.ac.uk}{1}
\addauthor{Triantafyllos Afouras}{afourast@robots.ox.ac.uk}{1}
\addauthor{Andrew Zisserman}{az@robots.ox.ac.uk}{1}

\addinstitution{
 Visual Geometry Group\\
 Department of Engineering Science\\
 University of Oxford\\
 Oxford, UK
}

\runninghead{Prajwal et al.}{Visual Keyword Spotting with Attention}

\def\eg{\emph{e.g}\bmvaOneDot}

\def\etal{\emph{et al}\bmvaOneDot}

\begin{document}

\maketitle

\blfootnote{$^\ast$ Equal Contribution.}

\begin{abstract}

In this paper, we consider the task of spotting spoken keywords in silent video sequences -- also known as visual keyword spotting.
To this end, we investigate 
 Transformer-based models that ingest two streams, a visual encoding of the video and a phonetic encoding of the keyword, and output the temporal location of the keyword if present. Our contributions are as follows: (1) We propose a novel architecture, the {\em Transpotter}, that uses full cross-modal attention between the visual and phonetic streams; (2) We show through extensive evaluations that our model outperforms the prior state-of-the-art visual keyword spotting and lip reading methods on the challenging LRW, LRS2, LRS3 datasets by a large margin; (3) We demonstrate the ability of our model to spot words under the extreme conditions of isolated mouthings in sign language videos.

\end{abstract}
\psec
\section{Introduction}\label{sec:intro}
\psec

In recent years, there has been significant progress in automatic visual speech recognition (VSR) due to the availability of large-scale annotated datasets and the development of powerful neural network-based learners~\cite{Chung17,Assael16,Afouras19}.
These methods are continually improving and becoming more sophisticated, by incorporating
better visual models, stronger language modelling and training on larger datasets.
Indeed the best industrial grade lip reading models today are far superior to humans, and achieve
error rates approaching Automatic Speech Recognition (ASR) performance~\cite{makino2019recurrent,prajwal2021sub}.

However, for many applications it is not necessary to transcribe every word that is spoken in a silent video (the task of VSR), rather only specific utterances or keywords need to be recognised.
This is for example the case in ``wake word'' recognition, where
only particular keywords need to be spotted over long input sequences.
A further drawback of VSR methods is that they are heavily reliant on language modelling; in general,
their performance decreases significantly when context is limited (e.g.\ short utterances) or parts of the input are occluded, e.g.\ from the speaker's hands or a microphone.
In this work, we focus instead on the task of \textit{Visual Keyword Spotting} (KWS), where the the goal is to detect and localise a \textit{given} keyword in (silent) spoken videos.

Automatic visual KWS enables a diverse range of practical applications: indexing archival silent videos by keyword to enable content-based search; helping virtual assistants (e.g.\ Alexa and Siri) and smart home technologies respond to wake words and phrases; assisting people with speech impairment (e.g.\ amyotrophic lateral sclerosis patients) or aphonia in communication~\cite{Shillingford18}; and detecting mouthings in sign language videos~\cite{Albanie20}.

KWS differs in complexity from VSR primarily because in KWS we are armed with the keyword we need to recognise, whereas VSR has the harder task of recognising every word from scratch.
The core hypothesis motivating this
work is that \textit{this additional knowledge renders visual KWS an easier task} than VSR; and it is therefore expected 
that KWS should achieve a higher performance than VSR, and generally be more robust to challenging and adversarial situations. 
Nevertheless, visual KWS remains a \textit{very difficult} task and shares similar challenges to VSR methods: first, some words sound different but involve identical lip movements (`man', `pan', `ban'), these {\it homopheme} words cannot be distinguished using only visual information. Second, speech variations such as accents, speed, and mumbling can alter lip movements significantly for the same word. Third, co-articulation of the lips between preceding and subsequent words in continuous speech also affects lip appearance and motion.

In this paper, we make the following three contributions: (i)~We propose a novel Transformer-based architecture, the {\em Transpotter} (a portmanteau of {\em Trans}former and {\em Spotter}), that is tailored to the visual KWS task. The model takes as input two streams, one encoding visual information from a video and the other providing a phonetic encoding of the keyword; the heterogeneous inputs are then fused using full cross-modal attention. (ii)~Through extensive evaluations, we show that our Transpotter model outperforms the prior state-of-the-art visual KWS and VSR methods on the challenging LRW, LRS2 and LRS3 lip reading datasets by a large margin. (iii)~We test our best model under extreme conditions: finding words in mouthings of people communicating using sign language. Signers sometimes mouth words as they sign as an additional non-manual signal to disambiguate and help understanding~\cite{sutton2007mouthings}. This new task is extremely challenging as there is a significant domain shift  between full spoken sentences (in our training and test sets) and mouthings, where the context is sporadic and phonemes of the keyword may be missing -- as sometimes only parts of words are mouthed~\cite{Boyes2001mouthing}. Our approach outperforms previous KWS models in this challenging, practical use-case.
Video examples are available at the project's webpage:~\url{www.robots.ox.ac.uk/~vgg/research/transpotter}.

\psec
\section{Related work}\label{sec:related_work}
\psec

Our work relates to prior work on KWS, lip reading, visual grounding, and applications of Transformers for text and video. 
We present a brief discussion of these topics below.  

\newpara\noindent{\textbf{KWS.}} 
KWS in audio (speech) is a well studied problem with a long history, spanning several decades. Prior to the establishment of deep learning models, KWS methods were based on Hidden Markov Models~\cite{rose1990hidden,Wilpon89}, dynamic time warping~\cite{Fumitada90dtw,Sakoe78dynamicprogramming,Yaodong10} or indexing of ASR lattices~\cite{Can2011LatticeIF}.
A number of works have since used deep architectures suitable for sequence modelling
(e.g. RNNs, CNNs, or graph convolutional networks)
\cite{Sainath2015ConvolutionalNN,HelloEdge,Wang2, Palaz,graph_cnn, Fernandez, HWang,
Sun2016MaxpoolingLT, Arik, Lengerich, Kim}, including encoder-decoder approaches
\cite{Zhang18,Audhkhasi,Zhuang2016UnrestrictedVK,rosenberg2017end}.
Berg~\etal~\cite{berg2021keyword} recently proposed using a Transformer model for the same task. Different from ours, this work uses a single input stream (audio) and only learns to spot a fixed vocabulary of keywords.
In contrast, we use Transformers to temporally process, then fuse the multi-modal inputs, building a model that can eventually perform open-set KWS.  
Visual KWS has also received attention recently.
The proposed methods include query-by-example~\cite{Jha} approaches, sliding window classification~\cite{sliding-windows}, or
looking up phonetic queries in lip reading feature sequences~\cite{stafylakis18kws, Momeni20}, while
audio-visual methods~\cite{Wu16avkws,Ding,Momeni20} that fuse the two
modalities to improve robustness to noise have also been proposed. 
Our method builds upon these approaches: we address various weaknesses and propose superior video-text modelling as well as explicit keyword localization, resulting in significantly improved performance.

\noindent{\textbf{Lip reading.}}
Early works in lip reading usually relied on hand-crafted pipelines and features~\cite{potamianos2003recent,gowdy2004dbn,papandreou09,Zhou14}.
The availability of large scale lip reading datasets~\cite{Chung17,Afouras18d} and the development
of deep neural network models resulted in major performance improvements, initially in 
word-level lip reading~\cite{Chung16,Stafylakis17} and constrained sentences~\cite{Assael16}.
Sentence level models were subsequently developed, using sequence-to-sequence
architectures based on RNNs~\cite{Chung17}, CTC-based~\cite{Shillingford18} approaches,
or a hybrid of the two \cite{petridis2018audio}.
Replacing RNNs with Transformers resulted in better performing
architectures~\cite{Afouras19,zhang2019spatio,gulati2020conformer}.
Joint audio-visual training and cross-modal distillation~\cite{Afouras20,Jianwei20tdnn,li2019improving} have also been investigated.
The current state-of-the-art model uses Transformers in the visual front-end and achieves remarkable results with word error rates reaching as low as 30.7\%~\cite{prajwal2021sub}. 

\newpara\noindent{\textbf{Visual grounding.}} 
Our work is also related to tasks such as
natural language grounding in 
videos~\cite{Hendricks2017LocalizingMI,gao2017tall,
Liu2018AttentiveMR,Xu2019MultilevelLA,Yuan2019ToFW,
Ghosh2019ExCLEC,Chen2018TemporallyGN,Zeng2020DenseRN}
and subtitle alignment in sign language clips~\cite{Bull21}.

\newpara\noindent{\textbf{Transformers.}} 
Since their introduction for machine translation, Transformers~\cite{vaswani2017attention} have become ubiquitous and are used today in a wide range of applications from natural language processing~\cite{devlin2018bert,radford2019language} and speech recognition~\cite{Dong2018SpeechTransformerAN,Karita2019ImprovingTE,mohamed2019transformers} to visual representation learning~\cite{dosovitskiy2021an,bertasius2021spacetime,wu20visual,prajwal2021sub}.
In this work, we rely on Transformers as our building blocks for their
strong sequence modelling capability and inherent potential for localisation through attention.

\psec
\section{Visual KWS with Attention}\label{sec:method}
\psec

\begin{figure}[t]
    \centering
    \includegraphics[width=\textwidth]{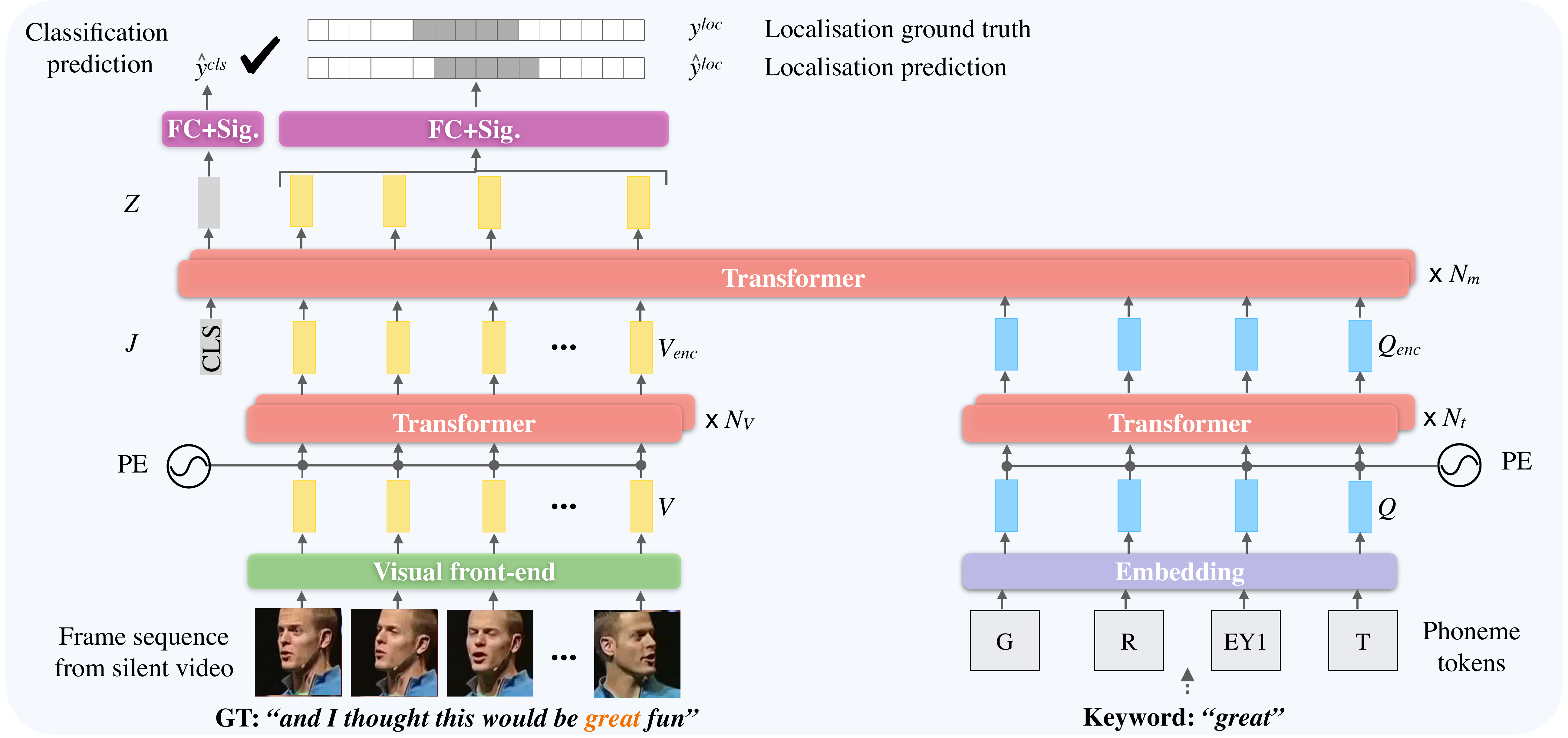}
    \caption{\textbf{The Transpotter architecture:} Video frames are inputted to a visual front-end (CNN~\cite{Afouras19} or VTP~\cite{prajwal2021sub}) to extract low-level visual features, which are then passed to $N_v$ Transformer layers to encode temporal information. The keyword in the form of a phoneme sequence is encoded using $N_t$ Transformer layers. The text and visual features are finally concatenated in time and processed using a joint multi-modal Transformer which predicts: (i) the probability the keyword occurs in the video, (ii) frame-level probabilities indicating the location of the word. PE corresponds to positional encoding.
    }
    \label{fig:architecture}
    \pfig
\end{figure}

In this section, we describe our proposed method shown in Figure~\ref{fig:architecture}.
We outline the architecture of our model (Section~\ref{subsec:architecture}), our training procedure (Section~\ref{subsec:training}) and differences to prior work (Section~\ref{subsec:discussion}). We refer the reader to \supp for further details.

\psubsec
\subsection{The Transpotter Architecture}\label{subsec:architecture}
\psubsec
Our model ingests two input streams: (i)~a textual keyword 
$q = (q_1, q_2 \cdots, q_{n_{p}})$, and (ii)~a silent video clip
$v \in \mathbb{R}^{T \times H \times W \times 3}$
in which we need to spot the keyword. 
For each of the inputs, we have separate encoders that learn initial modality-specific
representations.
This is followed by a joint multi-modal Transformer that learns cross-modal
relationships between the video and text features.
The joint transformer predicts two outputs: 
(i) a sequence-level probability of the keyword occurring in the video and
(ii) frame-level probabilities indicating the location of the 
keyword in the video if present. We describe each of the modules next. 

\noindent
\textbf{Text Representations.} Our textual input is a phonetic representation of the keyword, obtained using a pronunciation dictionary. 
The input phoneme sequence of length $n_p$ is mapped to a sequence 
of learnable embedding vectors 
$Q \in \mathbb{R}^{n_p \times d}$.
Sinusoidal positional encodings are added to the input phoneme feature vectors,
and the result is passed through a Transformer Encoder~\cite{vaswani2017attention} with
$N_t$ layers to capture temporal information across the phoneme sequence:
\vspace{-5pt}
\[
Q_{enc} =
encoder_q(Q + PE_{1:n_p}) \in \mathbb{R}^{n_p \times d}.
\]
\vspace{-5pt}

\noindent
\textbf{Video Representations.} We use a pre-trained visual front-end (either a CNN~\cite{Afouras19} or VTP~\cite{prajwal2021sub}) to
extract a feature vector for each input video frame,
$V \in \mathbb{R}^{T \times d}$.
Similar to the text encoder $Q_{enc}$, we pass the visual features through a Transformer Encoder with $N_v$ layers to capture temporal information, after adding positional encodings:
\vspace{-5pt}
\[
V_{enc} =
encoder_v(V + PE_{1:T}) \in \mathbb{R}^{T \times d}.
\]
\vspace{-5pt}

\noindent
\textbf{Joint Video-Text Representations.} 
The uni-modal representations $V$ and $Q$ are concatenated along the time dimension to produce a single sequence of feature vectors. A learnable $[CLS]$ token embedding (such as in BERT~\cite{devlin2018bert} and ViT~\cite{dosovitskiy2021an}) is then prepended to the result:
\[
J = ( [CLS]; V_{enc}; Q_{enc} ) \in \mathbb{R}^{ (1+ T + n_p)\times d}.
\]
We use a Transformer encoder with $N_m$ layers to jointly
learn the relationships across video and phoneme vectors:
\[
Z =
encoder_{vq} ( J + PE_{1:(1+T+n_p)}) \in \mathbb{R}^{(1+T) \times d}.
\footnote{the $n_p$ outputs corresponding to the phonetic embeddings are dropped.}
\]

\noindent
\textbf{Prediction heads.}
The $[CLS]$ output feature vector $Z_1$ serves as a joint aggregate
representation for the video-text pair. 
An MLP head for binary classification, $f_c$ is attached to $Z_1$ to predict the probability of the keyword being present in the video:
\vspace{-5pt}
\[
\hat{y}^{cls} = \sigma( f_{c} ( Z_1 ) ) \in \mathbb{R}^{1},
\]
where $\sigma$ denotes a sigmoid activation.
To localise the keyword, we attach a second MLP head $f_l$ that is shared across all the video output states from the multi-modal joint Transformer:
\[
\hat{y}^{loc} = \sigma( f_{l} ( Z_{2:(T+1)} ) ) \in \mathbb{R}^{T}.
\]
The output $y^{loc}_t$ at each video frame time-step $t \in T$ indicates the probability of the frame $t$ being a part of the keyword utterance.

\psubsec
\subsection{Training} \label{subsec:training}
\psubsec
\newpara\noindent{\textbf{Optimisation objectives.}}
Given a training dataset $\mathcal{D}$ consisting of tuples
$(v, q, y^{cls}, y^{loc})$ of silent video clips, text queries, class labels and location labels
(indicating the position of the keyword within the clip), we define the following objectives:

\vspace{-15pt}
\begin{align}
\mathcal{L}^{cls}
&=
-
\mathbb{E}_{(v, q, y^{cls}) \in \mathcal{D}} \
BCE(y^{cls},\hat{y}^{cls})
\label{eqn:loss_cls}
\\
\mathcal{L}^{loc}
&=
-
\mathbb{E}_{(v, q, y^{cls}, y^{loc}) \in \mathcal{D}} \
y^{cls}
\left[
\frac{1}{T}
\sum_{t=1}^{T}
BCE(y^{loc}_t ,\hat{y}^{loc}_t )
\right]
\label{eqn:loss_loc}
\\
BCE(y,\hat{y})
&=
y \log \hat{y} + (1-y) \log (1-\hat{y}),
\label{eqn:bce}
\end{align}
where BCE stands for the binary cross-entropy loss. The labels $y^{cls}$ are set to $1$ when the given keyword occurs in the video and 0 otherwise;
the frame labels $y^{loc}$ are set to 1 for the frames where the keyword is uttered and 0 otherwise.
We train the model the optimise the total loss
$
\mathcal{L}
=
\lambda \mathcal{L}^{cls} + (1-\lambda) \mathcal{L}^{loc}
\label{eqn:loss}
$,
where $\lambda$ is a balancing hyper-parameter.

\psubsec
\subsection{Discussion}\label{subsec:discussion}
\psubsec

Compared to prior approaches, the design of our model offers several important advantages.

\noindent
\textbf{Stronger Visual Representations.} Previous works~\cite{stafylakis18kws,Momeni20}
model temporal relationships between video frames using
RNNs. In contrast, we employ
Transformers~\cite{vaswani2017attention}, which are far more
effective in modeling temporal relationships
~\cite{al2019character,hochreiter2001gradient}.

\noindent
\textbf{Joint Video-text Modeling.}
Prior works such as KWS-Net~\cite{Momeni20} follow a late-fusion strategy.
In our model each frame-wise video feature 
can attend to any keyword token (phoneme) and vice-versa. The information exchange across the modalities occurs at every layer, without
restrictions on the receptive field for either modality.  

\noindent
\textbf{Stronger keyword localisation.}
Fine-grained localisation of the keyword in the video can be important for applications such as sign spotting~\cite{Albanie20}. Existing methods \cite{Momeni20, stafylakis18kws} ``weakly'' localise the keyword by taking the sequence-level prediction to be the maximum probability over all the video time-steps. We instead provide stronger frame-level supervision, by enforcing the model to predict the exact temporal extent of the keyword in the video.

\psubsec
\subsection{Implementation details}
\psubsec
\label{subsec:impl}
\noindent
\textbf{Pre-training the visual front-end.} We explore two different visual front-end architectures for the Transpotter: (1)~a CNN, highly similar in architecture to TM-seq2seq ~\cite{Afouras19} and (2)~VTP~\cite{prajwal2021sub}, the current state-of-the-art for lip reading (trained only on public data). Both models are trained end-to-end on two-word video clips of LRS2~\cite{Chung17} and LRS3~\cite{Afouras18d} for lip reading. We refer the reader to \supp for the exact CNN architecture and training hyper-parameters. We refer the reader to~\cite{prajwal2021sub} for architectural hyper-parameters and training protocols for VTP. We pre-compute the visual features for each backbone for both datasets and then train directly on them for faster training. All our models and ablations use the pre-trained CNN features, unless otherwise stated. 

\noindent \textbf{Sampling.} We form the training dataset $\mathcal{D}$ by randomly sampling with $50\%$ probability a positive or negative video clip $v$ for each query $q$. Each video $v$ contains word boundary annotations, which allows (i) performing data augmentation by randomly cropping video segments during training, and (ii) creating frame labels $y^{loc}$, as described in~\ref{subsec:training}.

\noindent
\textbf{Misc.}
The keyword $q$ is mapped to a phoneme sequence using the CMU dictionary \cite{cmu}; words not present in the dictionary are discarded from training $\mathcal{D}$.
We set $\lambda=0.5$.

\psec
\section{Experiments}\label{sec:experiments}
\psec

This section is structured as follows: We first present the datasets used as well the evaluation protocols that we follow in our experiments (Section~\ref{subsec:datasets}). Next, we compare the performance of our proposed Transpotter model against strong baselines (Section~\ref{subsec:baselines}) and then present a comprehensive study ablating our design choices (Section~\ref{subsec:ablations}). Finally, we perform further performance analysis and provide qualitative results (Section~\ref{subsec:analysis}).

\psubsec
\subsection{Datasets and Evaluation Protocol}\label{subsec:datasets}
\psubsec

\noindent\textbf{Datasets.} 
All models and baselines are trained and evaluated on  
LRS2~\cite{Chung17} and LRS3~\cite{Afouras18d} lip reading datasets.
LRS2 contains BBC broadcast footage from British television and LRS3 is based on TED/TEDx videos downloaded from YouTube (refer to \supp for detailed statistics).
The video clips for both datasets are tightly cropped face-tracks of active speakers only.
For each clip, a full transcription of the utterance as well as
word boundary alignments are provided. The number of videos, number of keyword instances and keyword vocabulary for each of the test sets is shown in Table~\ref{tab:baselines}.

\noindent\textbf{Evaluation Protocol.}
Evaluation is performed for every test dataset as follows: 
First, the vocabulary of test keywords is determined, by considering all the words occurring in the test set transcriptions with above a certain phoneme length $n_p$. If not specified, we use $n_p\geq3$. Every word in the query vocabulary is then searched for in all the test set videos. 

\noindent\textbf{Metrics.}
Given ground truth video-keyword samples, we assess the performance of our model in two ways.  
First, we assess classification performance, \ie~whether the model can accurately predict whether the keyword occurs in the video or not. We compute accuracy (\ackcls) and mean average precision (\mapcls) metrics, where \ackcls~measures how often a given keyword occurs in any of the top-$k$ retrievals, and \mapcls~is obtained with the above criterion (where every word in the test keyword vocabulary is considered as a separate class).
 
Second, we assess the model's localisation capability, \ie~whether the model can accurately localise the keyword in the video clip. We follow common practice from the detection literature: we consider a keyword accurately detected when the intersection-over-union (IOU) between the prediction $\hat{y}^{loc}$ and ground truth label ${y}^{loc}$ is above a certain threshold, and calculate the mean average precision \maploc.
To calculate the IOU, we binarise the model's predictions using a threshold $\tau=0.5$.

\psubsec
\subsection{Comparison to baselines}\label{subsec:baselines}
\psubsec
We compare our model's performance against a state-of-the-art VSR model and KWS-Net~\cite{Momeni20}, the previous state-of-the-art visual KWS model. 

\noindent
\textbf{VSR baseline.} We use an improved version of the TM-seq2seq~\cite{Afouras19} VSR model, with the same pre-trained CNN backbone (Section~\ref{subsec:impl}) that we use for the KWS models. The model is trained with the curriculum training strategy of~\cite{Afouras19} (details in \supp). The VSR model achieves state-of-the-art Word Error Rate (WER) performance of $36.9$\% and $48.0$\% on the LRS2, LRS3 test sets respectively.
Since the VSR model only produces text transcriptions of a given video, but no localisation prediction, we can only evaluate its classification performance (\ackcls,\mapcls).
We follow the method detailed in~\cite{he} to estimate the posterior probability that the keyword occurs in a video clip.

\noindent
\textbf{KWS-Net.} As a KWS baseline we use the state-of-the-art model of Momeni \etal~\cite{Momeni20}. For fair comparison, here too we use the same CNN backbone that is also used for our model.

\begin{table}[ht]
    \setlength{\tabcolsep}{8pt}
    \centering
    \resizebox{0.99\linewidth}{!}{
        \begin{tabular}{l|ccc|c|ccc|c}
            \toprule
             & \multicolumn{4}{c}{LRS2} & \multicolumn{4}{|c}{LRS3}\\
            \midrule
            & \multicolumn{4}{c}{ 1.2K vids. / 4.3K inst. / 1.6K vocab.} 
            & \multicolumn{4}{|c}{1.3K vids. / 6.1K inst. / 1.9K vocab.} \\
            \midrule
            Model  & \aconecls & \acfivecls  & \mapcls  &\maploc & \aconecls & \acfivecls  & \mapcls   &\maploc \\
            \midrule
            KWS-Net~\cite{Momeni20}  &  36.1 & 61.2 & 41.0& 36.2  & 29.8 & 54.6 &34.3 & 29.2 \\
            VSR &  63.7 & 76.3 & 64.3 & - & 52.3 & 66.0 & 50.3 & - \\
            Transpotter  &  65.0 & 87.1 & 69.2 &68.3  & 52.0 & 77.1 & 55.4  & 53.6 \\
            \midrule
            Transpotter (VTP)  &\textbf{ 68.7 }& \textbf{90.7 }&\textbf{72.5}&\textbf{ 71.6}&\textbf{ 55.7 } &\textbf{ 78.5}& \textbf{58.2} &\textbf{ 56.1}  \\
            \bottomrule
        \end{tabular}
    }
    \vspace{1pt}
    \caption{\textbf{Comparison to baselines:} We outperform the current state-of-the-art KWS and VSR methods by a large margin. Our Transpotter model is particularly effective in localising the keyword in the video. Moreover, by using the recently proposed VTP~\cite{prajwal2021sub} architecture as the Transpotter's visual backbone instead of a CNN, we  achieve even better performance.
    }
    \label{tab:baselines}
    \ptab
    \vspace{-10pt}
\end{table}

\noindent
\textbf{State-of-the-art KWS.}
We report our model's performance and compare it with strong baselines in Table~\ref{tab:baselines}.
It is clear that our model outperforms both baselines. On the last row, we show the boost in performance by replacing the CNN with the recently proposed VTP backbone~\cite{prajwal2021sub}, resulting in state-of-the art performance on both the LRS2 and LRS3 datasets.

\noindent
\rebuttal{\textbf{Evaluation on LRW.}
We also compare the performance of KWS-Net~\cite{Momeni20} with our proposed Transpotter model on the LRW~\cite{Chung16} test set following the same evaluation protocol. The test set contains $25K$ single-word video clips spanning a vocabulary of $500$ words ($50$ instances per word). Note that KWS-Net has been pretrained on the LRW training split, but the Transpotter has only been trained on LRS2 and LRS3. As we can see in Table~\ref{tab:lrw_results}, the Transpotter outperforms the previous state-of-the-art baseline KWS-Net by a large margin.
We refer the reader to \supp for a qualitative error analysis in this setting.
}

\begin{table}[ht]
    \setlength{\tabcolsep}{30pt}
    \centering
    \resizebox{0.999\linewidth}{!}{
        \begin{tabular}{l|ccc}
            \toprule
            Model & \aconecls & \acfivecls  & \mapcls   \\
            \midrule
            KWS-Net~\cite{Momeni20} & 66.6 & 89.0  & 33.0   \\
          Transpotter   & \textbf{85.8} & \textbf{99.6} & \textbf{64.1} \\
            \bottomrule
        \end{tabular}
    }
    \vspace{-5pt}
    \caption{\textbf{Comparison on LRW~\cite{Chung16}:} The Transpotter outperforms the previous state-of-the-art KWS model on the LRW test set, despite not having been trained on LRW data.
    The localization metric~\maploc~is not reported as the input videos are single-word clips. 
    }
    
    \label{tab:lrw_results}
    \ptab
    \vspace{-15pt}
\end{table}

\psubsec
\subsection{Architecture ablations}\label{subsec:ablations}
\psubsec

To assess our design choices for the Transformer skeleton,
we perform a number of ablations considering variations of the model architecture. 
We briefly explain the alternative approaches below; more details can be found in \supp.

In particular we consider two alternative encoder-decoder architectures, 
with the video input at the encoder side and the text query at the decoder ($\text{Enc}_{vid}$-$\text{Dec}_{text}$) and vice versa ($\text{Enc}_{text}$-$\text{Dec}_{vid}$).
Since the latter model outputs at the temporal resolution of the video input, it can explicitly localise the keyword (in the same way as the Transpotter), while the former can only perform classification.
We also consider a variant of the Transpotter, where the model does not output localisation predictions (hence no $\mathcal{L}^{loc}$ is used for its training).
We show the results in Table~\ref{tab:model_ablations}.
The selected Transpotter architecture outperforms all variants. 
In particular, by comparing rows 2 and 4, we observe that training with a localisation head and loss $\mathcal{L}^{loc}$ also improves classification (e.g. $64.0$ vs $69.2$ \mapcls).

\begin{table}[ht]
    \setlength{\tabcolsep}{6pt}
    \centering
    \resizebox{0.999\linewidth}{!}{
        \begin{tabular}{l|ccc|c|ccc|c}
            \toprule
           &   \multicolumn{4}{c}{LRS2} & \multicolumn{4}{|c}{LRS3}\\
            \midrule
             Model  & \aconecls & \acfivecls  & \mapcls &\maploc & \aconecls & \acfivecls  & \mapcls  &\maploc \\
            \midrule
          $\text{Enc}_{vid}$-$\text{Dec}_{text}$  &  52.5 & 80.0 & 57.9 & -  &  40.3 & 66.9 & 43.2 & - \\
           Transpotter w/o loc. &  59.4 &84.1 & 64.0  & -  &  46.5 & 72.1 & 49.8 & - \\
           \midrule
            $\text{Enc}_{text}$-$\text{Dec}_{vid}$  &  63.8 & 86.8 & 68.4 & 67.8 &  \textbf{52.1} & 76.6 & 54.9 & 53.1 \\
           Transpotter  &  \textbf{65.0} & \textbf{87.1} &\textbf{ 69.2}  &\textbf{ 68.3 } &  52.0 & \textbf{77.1} & \textbf{55.4} & \textbf{53.6} \\
            \bottomrule
        \end{tabular}
    }
    \vspace{0.5pt}
    \vspace{-10pt}
    \caption{\textbf{Model ablations:} Our approach of jointly modeling text and video sequences with a localisation head for stronger supervision outperforms other architectural designs.
    }
    \label{tab:model_ablations}
    \ptab
\end{table}

\psubsec
\subsection{Transpotter performance analysis}\label{subsec:analysis}
\psubsec

\setlength{\tabcolsep}{1pt}
\begin{figure}[t]
    \centering
    \begin{tabular}{llll}
    (a) & \raisebox{-.5\height}{\includegraphics[width=0.45\textwidth]{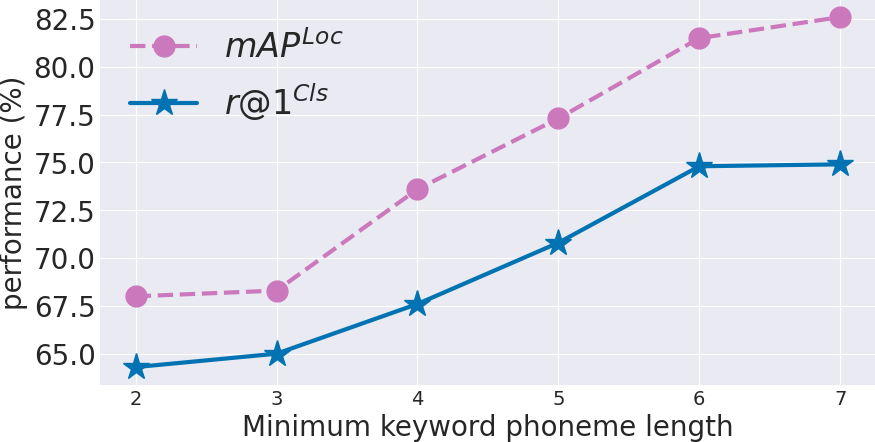}} & (b) & \raisebox{-.5\height}{\includegraphics[width=0.45\textwidth]{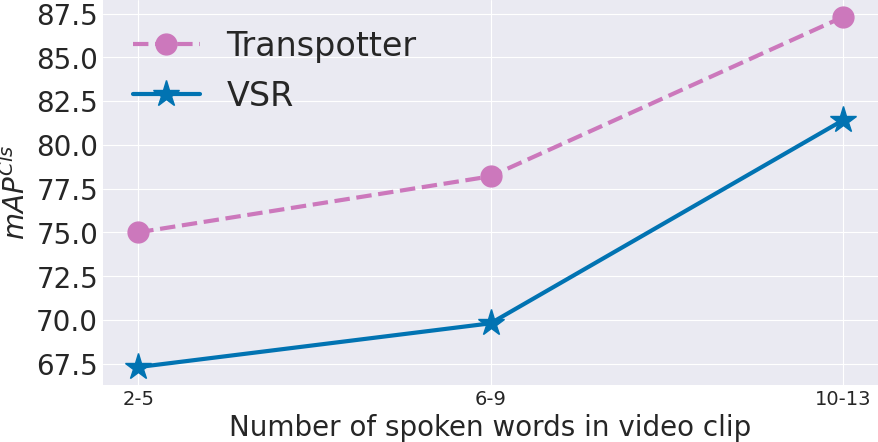}} %
    \end{tabular}
\caption{(a) Transpotter's performance increases with the keyword length; (b) Transpotter performs far better than VSR with limited context. Both methods improve with more context.}

  \label{fig:perf_analysis}
  \vspace{-5pt}
\end{figure}

In this section, we analyse the performance of our proposed method when varying the keyword length and the size of the surrounding visual context.

\noindent
\textbf{Keyword length.}
In Figure~\ref{fig:perf_analysis}a, we plot the model's performance 
on the LRS2 test set against the minimum keyword length in phonemes $n_p$.
As expected, longer keywords are easier to spot and therefore result in better retrieval performance. Indeed for long 7-phoneme keywords, \maploc~reaches as high as 82.5. We note however that even for very challenging short keywords with only 2 phonemes (such as "my", "to", "at"), \maploc~stays high at 67.5. 

\noindent
\textbf{Context.}
The visual appearance of spoken words can be highly ambiguous~\cite{Afouras19}, therefore recognising isolated words from visual input alone may be very challenging. Current lip reading  models utilise the surrounding visual context to resolve this ambiguity. In Figure~\ref{fig:perf_analysis}b, we illustrate how the performances of our Transpotter KWS model and our VSR baseline vary based on the amount of contextual information available. We plot the \mapcls~against the number of words in the video clip. We observe that both models benefit from larger surrounding context, with the Transpotter outperforming the VSR baseline consistently. %

\begin{figure}[t]
    \centering
    \includegraphics[height=180pt,width=\textwidth]{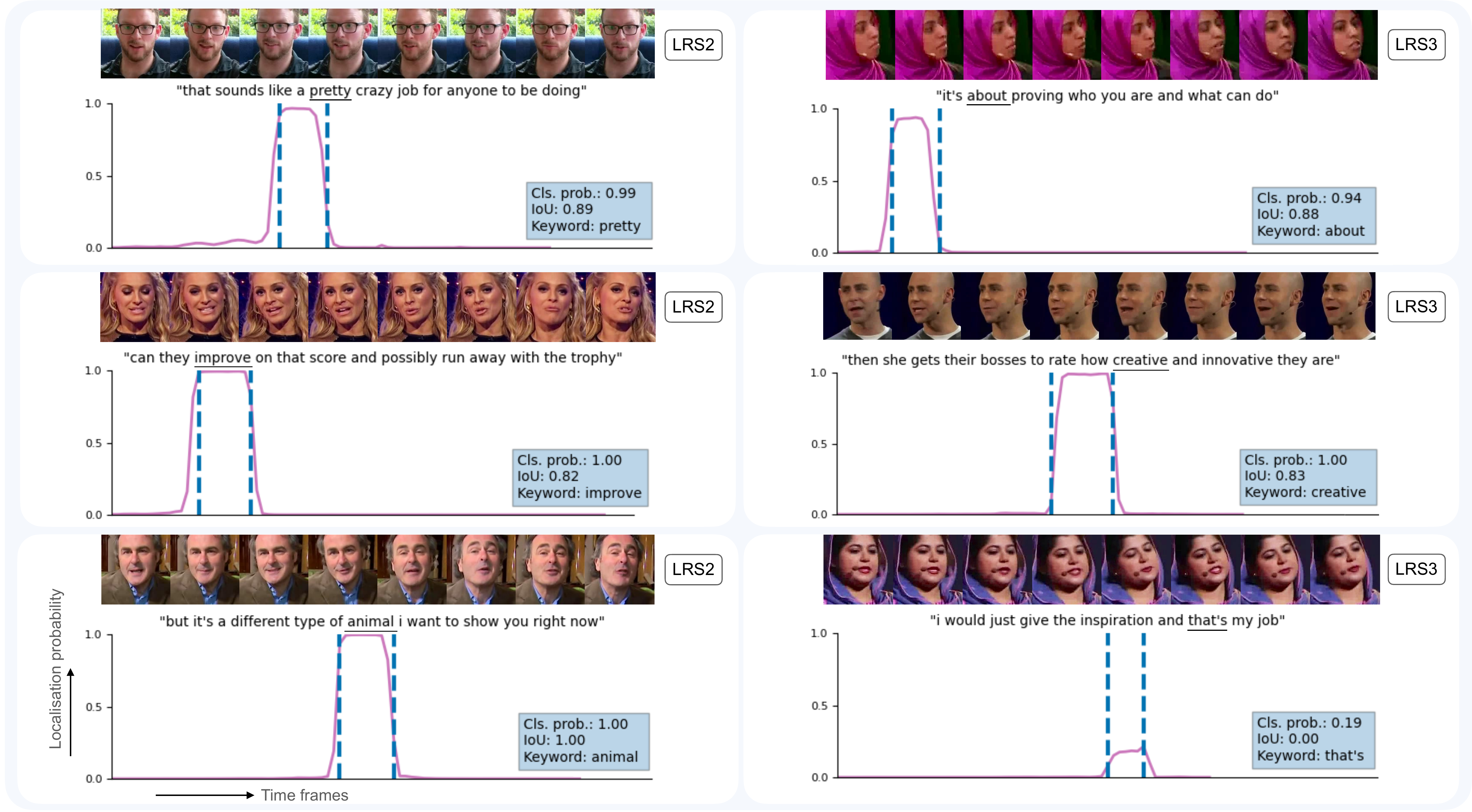}
    \vspace{-15pt}
    \caption{\textbf{Qualitative results on LRS2 and LRS3:} The Transpotter accurately localises the keyword in most examples. In the bottom right example, the model's confidence is low, most likely because it is a short word. The IOU is zero since we threshold at $\tau = 0.5$.}
    \label{fig:qualitative}
    \pfig
    \vspace{10pt}
\end{figure}

\noindent
\textbf{Qualitative analysis.}
In Figure~\ref{fig:qualitative}, we show qualitative examples from the LRS2 and LRS3 test sets. It is clear that the model produces smooth predictions that precisely indicate the full location of the word. In the bottom right corner we observe a failure case where the model's confidence is low -- the keyword ``that's'' in this case is short. 

\begin{figure}[t]
    \centering
    \includegraphics[width=\textwidth]{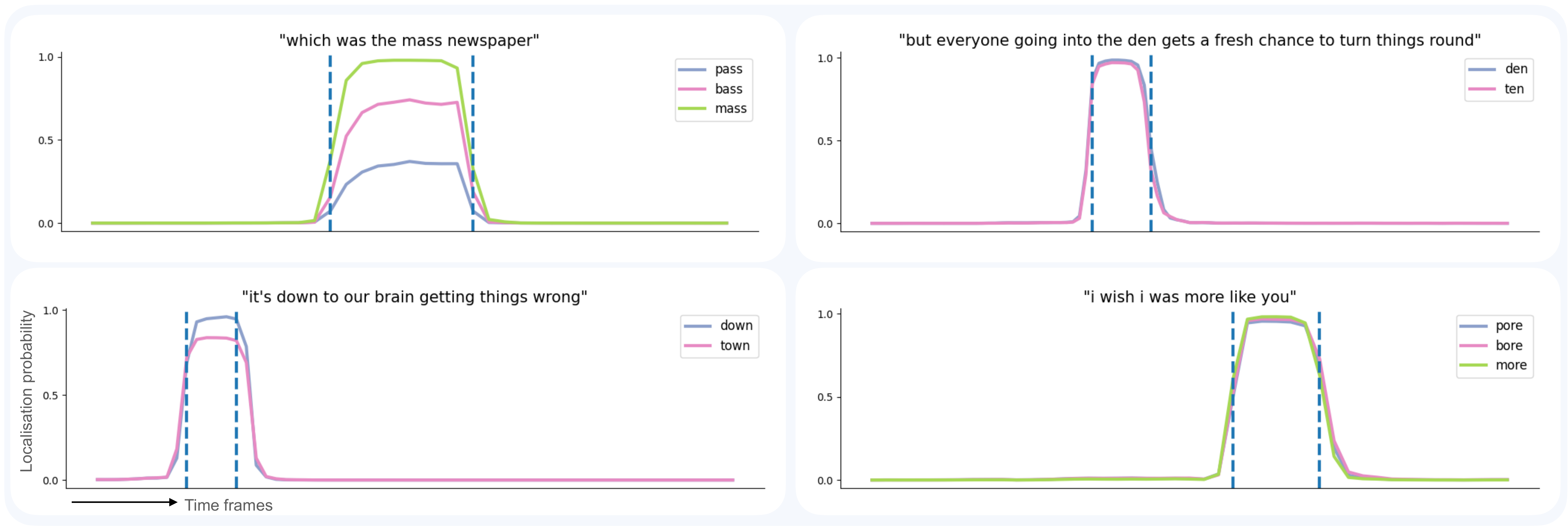}
    \caption{
    \textbf{Model's response to homophemes:} We query words and their corresponding homophemes for LRS2 test set clips.
    We observe that the model spots the words and their  homophemes  at the same (ground truth) location.
    }
    \label{fig:homophemes}
    \pfig
\end{figure}

\noindent
\textbf{Model  response to homophemes.}
We further probe our Transpotter model for failure cases. In visual-only keyword spotting, a common failure case is due to homophemes,
\ie~words with identical lip movements. 
To investigate the response of our model to such cases, we construct a list of keywords from the LRS2 test set sentences that are known to have homopheme counterparts (\eg \textit{mark}, which has two matching homophemes, \textit{bark} and \textit{park}) and then for each test set clip that contains one of the keywords, we query that keyword along with its corresponding homophemes and plot the model's outputs.
We illustrate several examples in Figure~\ref{fig:homophemes}. We observe that in such cases, the model spots the keyword as well as its homophemes  at the same (ground truth) location.

\psec
\section{Mouthing Spotting in Sign Language videos}\label{sec:applictions}
\psec

In this section, we investigate the application of our method for spotting mouthed words in sign language videos.
This is an important application of visual KWS, as it has enabled an entire line of work on sign language recognition~\cite{Albanie20,Momeni20b,varol21}.

\noindent\textbf{Data description \& evaluation protocol.}
Here, we use a subset of BSL Corpus~\cite{schembri2013building,bslcorpus17} as a test set.
BSL Corpus is a large public dataset containing videos of conversations conducted in sign language by deaf signers, from various regions across the UK. 
We extend the dataset's annotations by adding a \textit{Mouthing} tier and asking a deaf annotator to identify and localise mouthing occurrences that correspond to visible signs. We obtain $383$ mouthing instances, from $29$ different signers, over a keyword vocabulary size of $187$.
We use a pre-processing pipeline similar to~\cite{Chung17} to obtain face-cropped tracks around the faces of the signers.   
To evaluate KWS performance, we take 8-second video clips centered around the annotated mouthings and follow the same evaluation protocol described in Section~\ref{sec:experiments}.

\begin{figure}[ht]
    \centering
    \includegraphics[height=200pt,width=\textwidth]{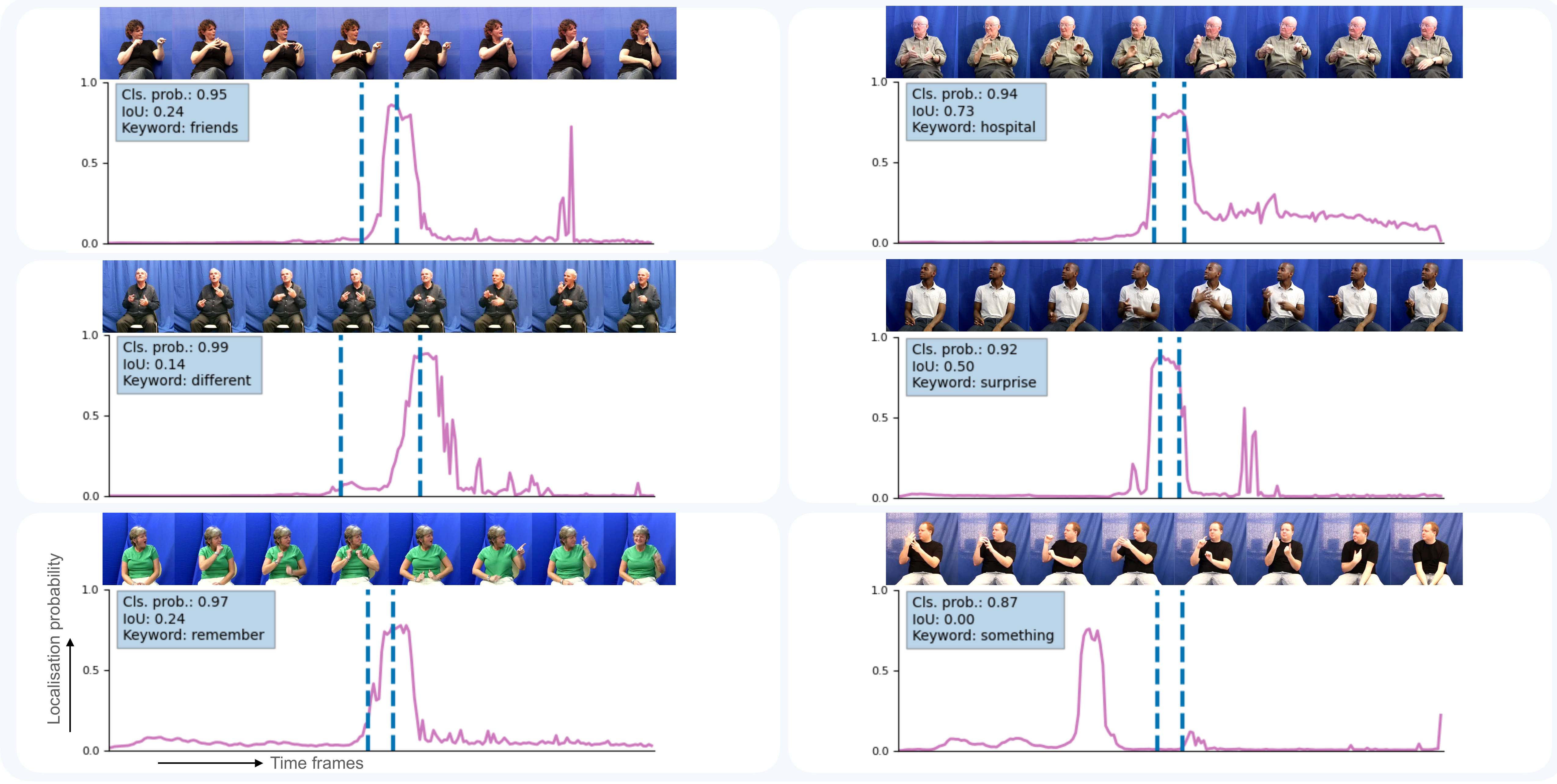}
    \caption{\textbf{Qualitative results on BSL-Corpus:} Despite the large
    domain shift from our training examples and additional challenges such as
    partially mouthed words and hand occlusions, 
    the Transpotter succeeds in correctly spotting mouthings in these
    challenging conditions. We observe a failure case (bottom right) where the localisation is incorrect.
    We note that contrary to LRS2 and LRS3, where word boundaries are obtained through robust audio-based forced alignment, the annotations for BSL-Corpus are noisier as they are performed manually.
    }
    \label{fig:qualitative_sl}
    \vspace{-5pt}
\end{figure}

\noindent\textbf{Results.}
We summarise the evaluation results in Table~\ref{tab:sign_lang}. The Transpotter model is far superior to the prior state-of-the-art KWS baseline, achieving a great improvement in performance (e.g.\ $29.6$ vs $15.6$ \mapcls~score).
To complete this analysis, we also show qualitative examples of the spotted mouthings in Figure~\ref{fig:qualitative_sl}. 

\noindent\textbf{Discussion.} 
We note that sign language mouthings are often very different from equivalent spoken words. Words may be partially mouthed and can be occluded by the signing hands. There is therefore a significant domain gap between the BSL-Corpus signing videos and our lip reading training videos.
However, we note that our proposed model greatly outperforms the KWS-Net baseline -- a variant of which has been successfully deployed for detecting mouthings in order to bootstrap learning of sign spotting methods~\cite{Albanie20,Momeni20b,varol21}.
This indicates the potential of our proposed method to greatly improve these pipelines.

\begin{table}[ht]
    \setlength{\tabcolsep}{30pt}
    \centering
    \resizebox{0.999\linewidth}{!}{
        \begin{tabular}{l|ccc}
            \toprule
             Model & \aconecls & \acfivecls  & \mapcls   \\
            \midrule
            KWS-Net~\cite{Momeni20} & 12.4 & 29.6 & 15.6  \\
           Transpotter    &  \textbf{22.5} & \textbf{47.1} & \textbf{29.6} \\
            \bottomrule
        \end{tabular}
    }
    \vspace{2pt}
    \caption{\textbf{Spotting mouthings in BSL-Corpus:} The Transpotter is far more accurate than the current state-of-the-art in spotting keywords in videos. 
    }
    \label{tab:sign_lang}
    \ptab
\end{table}

\psec
\psec
\section{Conclusion}\label{sec:conclusion}
\psec

We have presented the \textit{Transpotter}, a cross-modal attention based architecture for visual keyword spotting. Our method surpasses the performance of the previous best visual keyword spotting approach by a large margin, as well as that of a state-of-the-art lip reading baseline. We demonstrate the ability of our model to generalise to sign language videos where it can be used to spot mouthings, enabling automatic annotation of sign instances. In future work, we plan to further improve our method's performance by incorporating keyword semantics and context of the surrounding words.

\paragraph{Acknowledgements.} 
Funding for this research is provided by the UK EPSRC CDT in Autonomous Intelligent
Machines and Systems, the Oxford-Google DeepMind Graduate Scholarship, the EPSRC
Programme Grant VisualAI (EP/T028572/1) and the Royal Society Research Professorships 2019 RP/R1/191132. We thank Samuel Albanie for his invaluable help in applying our method to signer mouthings.

\bibliography{shortstrings,vgg_local,vgg_other,refs}

\begin{thebibliography}{78}
\providecommand{\natexlab}[1]{#1}
\providecommand{\url}[1]{\texttt{#1}}
\expandafter\ifx\csname urlstyle\endcsname\relax
  \providecommand{\doi}[1]{doi: #1}\else
  \providecommand{\doi}{doi: \begingroup \urlstyle{rm}\Url}\fi

\bibitem[Afouras et~al.(2018)Afouras, Chung, and Zisserman]{Afouras18d}
Triantafyllos Afouras, Joon~Son Chung, and Andrew Zisserman.
\newblock {LRS3-TED}: a large-scale dataset for visual speech recognition.
\newblock In \emph{arXiv preprint arXiv:1809.00496}, 2018.

\bibitem[Afouras et~al.(2019)Afouras, Chung, Senior, Vinyals, and
  Zisserman]{Afouras19}
Triantafyllos Afouras, Joon~Son Chung, Andrew Senior, Oriol Vinyals, and Andrew
  Zisserman.
\newblock Deep audio-visual speech recognition.
\newblock \emph{IEEE PAMI}, 2019.

\bibitem[Afouras et~al.(2020)Afouras, Chung, and Zisserman]{Afouras20}
Triantafyllos Afouras, Joon~Son Chung, and Andrew Zisserman.
\newblock Asr is all you need: Cross-modal distillation for lip reading.
\newblock In \emph{International Conference on Acoustics, Speech, and Signal
  Processing}, 2020.

\bibitem[Al-Rfou et~al.(2019)Al-Rfou, Choe, Constant, Guo, and
  Jones]{al2019character}
Rami Al-Rfou, Dokook Choe, Noah Constant, Mandy Guo, and Llion Jones.
\newblock Character-level language modeling with deeper self-attention.
\newblock In \emph{Proceedings of the AAAI Conference on Artificial
  Intelligence}, volume~33, pages 3159--3166, 2019.

\bibitem[Albanie et~al.(2020)Albanie, Varol, Momeni, Afouras, Chung, Fox, and
  Zisserman]{Albanie20}
Samuel Albanie, G{\"u}l Varol, Liliane Momeni, Triantafyllos Afouras, Joon~Son
  Chung, Neil Fox, and Andrew Zisserman.
\newblock {BSL-1K}: {S}caling up co-articulated sign language recognition using
  mouthing cues.
\newblock In \emph{ECCV}, 2020.

\bibitem[Arik et~al.(2017)Arik, Kliegl, Child, Hestness, Gibiansky, Fougner,
  Prenger, and Coates]{Arik}
Sercan Arik, Markus Kliegl, Rewon Child, Joel Hestness, Andrew Gibiansky, Chris
  Fougner, Ryan Prenger, and Adam Coates.
\newblock Convolutional recurrent neural networks for small-footprint keyword
  spotting.
\newblock In \emph{INTERSPEECH}, 2017.

\bibitem[Assael et~al.(2016)Assael, Shillingford, Whiteson, and
  de~Freitas]{Assael16}
Yannis~M. Assael, Brendan Shillingford, Shimon Whiteson, and Nando de~Freitas.
\newblock Lipnet: Sentence-level lipreading.
\newblock \emph{arXiv:1611.01599}, 2016.

\bibitem[Audhkhasi et~al.(2017)Audhkhasi, Rosenberg, Sethy, Ramabhadran, and
  Kingsbury]{Audhkhasi}
Kartik Audhkhasi, Andrew Rosenberg, Abhinav Sethy, Bhuvana Ramabhadran, and
  Brian Kingsbury.
\newblock End to-end asr-free keyword search from speech.
\newblock \emph{IEEE Journal of Selected Topics in Signal Processing}, 2017.

\bibitem[Berg et~al.(2021)Berg, O'Connor, and Cruz]{berg2021keyword}
Axel Berg, Mark O'Connor, and Miguel~Tairum Cruz.
\newblock Keyword transformer: A self-attention model for keyword spotting.
\newblock \emph{arXiv preprint arXiv:2104.00769}, 2021.

\bibitem[Bertasius et~al.(2021)Bertasius, Wang, and
  Torresani]{bertasius2021spacetime}
Gedas Bertasius, Heng Wang, and Lorenzo Torresani.
\newblock Is space-time attention all you need for video understanding?
\newblock In \emph{Proc. ICML}, 2021.

\bibitem[{Boyes Braem} and Sutton-Spence(2001)]{Boyes2001mouthing}
P~{Boyes Braem} and RL~Sutton-Spence.
\newblock \emph{The Hands Are The Head of The Mouth. The Mouth as Articulator
  in Sign Languages}.
\newblock Hamburg: Signum Press, 2001.
\newblock ISBN 3927731838.

\bibitem[Bull et~al.(2021)Bull, Afouras, Varol, Albanie, Momeni, and
  Zisserman]{Bull21}
Hannah Bull, Triantafyllos Afouras, G{\"u}l Varol, Samuel Albanie, Liliane
  Momeni, and Andrew Zisserman.
\newblock Aligning subtitles in sign language videos.
\newblock In \emph{arxiv}, 2021.

\bibitem[Can and Saraçlar(2011)]{Can2011LatticeIF}
Dogan Can and M.~Saraçlar.
\newblock Lattice indexing for spoken term detection.
\newblock \emph{IEEE Transactions on Audio, Speech, and Language Processing},
  19:\penalty0 2338--2347, 2011.

\bibitem[Chen et~al.(2018)Chen, Chen, Ma, Jie, and Chua]{Chen2018TemporallyGN}
Jingyuan Chen, Xinpeng Chen, L.~Ma, Zequn Jie, and Tat-Seng Chua.
\newblock Temporally grounding natural sentence in video.
\newblock In \emph{EMNLP}, 2018.

\bibitem[Chen et~al.(2019)Chen, Yin, Song, Ouyang, Liu, and Wei]{graph_cnn}
Xi~Chen, Shouyi Yin, Dandan Song, Peng Ouyang, Leibo Liu, and Shaojun Wei.
\newblock Small-footprint keyword spotting with graph convolutional network.
\newblock In \emph{IEEE Automatic Speech Recognition and Understanding
  Workshop}, 2019.

\bibitem[Chung and Zisserman(2016{\natexlab{a}})]{Chung16}
Joon~Son Chung and Andrew Zisserman.
\newblock Lip reading in the wild.
\newblock In \emph{Proc. ACCV}, 2016{\natexlab{a}}.

\bibitem[Chung and Zisserman(2016{\natexlab{b}})]{Chung16b}
Joon~Son Chung and Andrew Zisserman.
\newblock Signs in time: Encoding human motion as a temporal image.
\newblock In \emph{Workshop on Brave New Ideas for Motion Representations,
  ECCV}, 2016{\natexlab{b}}.

\bibitem[Chung et~al.(2017)Chung, Senior, Vinyals, and Zisserman]{Chung17}
Joon~Son Chung, Andrew Senior, Oriol Vinyals, and Andrew Zisserman.
\newblock Lip reading sentences in the wild.
\newblock In \emph{Proc. CVPR}, 2017.

\bibitem[Devlin et~al.(2018)Devlin, Chang, Lee, and Toutanova]{devlin2018bert}
Jacob Devlin, Ming{-}Wei Chang, Kenton Lee, and Kristina Toutanova.
\newblock {BERT:} pre-training of deep bidirectional transformers for language
  understanding.
\newblock \emph{CoRR}, abs/1810.04805, 2018.

\bibitem[Ding et~al.(2018)Ding, Pang, and Liu]{Ding}
Runwei Ding, Cheng Pang, and Hong Liu.
\newblock Audio-visual keyword spotting based on multidimensional convolutional
  neural network.
\newblock In \emph{IEEE International Conference on Image Processing}, 2018.

\bibitem[Dong et~al.(2018)Dong, Xu, and Xu]{Dong2018SpeechTransformerAN}
Linhao Dong, Shuang Xu, and Bo~Xu.
\newblock Speech-transformer: A no-recurrence sequence-to-sequence model for
  speech recognition.
\newblock \emph{2018 IEEE International Conference on Acoustics, Speech and
  Signal Processing (ICASSP)}, pages 5884--5888, 2018.

\bibitem[Dosovitskiy et~al.(2021)Dosovitskiy, Beyer, Kolesnikov, Weissenborn,
  Zhai, Unterthiner, Dehghani, Minderer, Heigold, Gelly, Uszkoreit, and
  Houlsby]{dosovitskiy2021an}
Alexey Dosovitskiy, Lucas Beyer, Alexander Kolesnikov, Dirk Weissenborn,
  Xiaohua Zhai, Thomas Unterthiner, Mostafa Dehghani, Matthias Minderer, Georg
  Heigold, Sylvain Gelly, Jakob Uszkoreit, and Neil Houlsby.
\newblock An image is worth 16x16 words: Transformers for image recognition at
  scale.
\newblock In \emph{International Conference on Learning Representations}, 2021.
\newblock URL \url{https://openreview.net/forum?id=YicbFdNTTy}.

\bibitem[Fernandez et~al.(2007)Fernandez, Graves, and Schmidhuber]{Fernandez}
Santiago Fernandez, Alex Graves, and Jurgen Schmidhuber.
\newblock An application of recurrent neural networks to discriminative keyword
  spotting.
\newblock In \emph{International Conference on Artificial Neural Networks},
  2007.

\bibitem[Gao et~al.(2017)Gao, Sun, Yang, and Nevatia]{gao2017tall}
Jiyang Gao, Chen Sun, Zhenheng Yang, and Ram Nevatia.
\newblock Tall: Temporal activity localization via language query.
\newblock In \emph{ICCV}, 2017.

\bibitem[Ghosh et~al.(2019)Ghosh, Agarwal, Parekh, and
  Hauptmann]{Ghosh2019ExCLEC}
Soham Ghosh, Anuva Agarwal, Zarana Parekh, and Alexander Hauptmann.
\newblock Excl: Extractive clip localization using natural language
  descriptions.
\newblock In \emph{NAACL-HLT}, 2019.

\bibitem[Gowdy et~al.(2004)Gowdy, Subramanya, Bartels, and
  Bilmes]{gowdy2004dbn}
John~N Gowdy, Amarnag Subramanya, Chris Bartels, and Jeff Bilmes.
\newblock Dbn based multi-stream models for audio-visual speech recognition.
\newblock In \emph{2004 IEEE International conference on acoustics, speech, and
  signal processing}, volume~1, pages I--993. IEEE, 2004.

\bibitem[Gulati et~al.(2020)Gulati, Qin, Chiu, Parmar, Zhang, Yu, Han, Wang,
  Zhang, Wu, and Pang]{gulati2020conformer}
Anmol Gulati, James Qin, Chung-Cheng Chiu, Niki Parmar, Yu~Zhang, Jiahui Yu,
  Wei Han, Shibo Wang, Zhengdong Zhang, Yonghui Wu, and Ruoming Pang.
\newblock Conformer: Convolution-augmented transformer for speech recognition.
\newblock In \emph{INTERSPEECH}, 2020.

\bibitem[He et~al.(2017)He, Prabhavalkar, Rao, Li, Bakhtin, and McGraw]{he}
Yanzhang He, Rohit Prabhavalkar, Kanishka Rao, Wei Li, Anton Bakhtin, and Ian
  McGraw.
\newblock Streaming small-footprint keyword spotting using sequence-to-sequence
  models.
\newblock In \emph{Proceedings of IEEE ASRU}, 2017.

\bibitem[Hendricks et~al.(2017)Hendricks, Wang, Shechtman, Sivic, Darrell, and
  Russell]{Hendricks2017LocalizingMI}
Lisa~Anne Hendricks, O.~Wang, E.~Shechtman, Josef Sivic, Trevor Darrell, and
  Bryan~C. Russell.
\newblock Localizing moments in video with natural language.
\newblock In \emph{ICCV}, 2017.

\bibitem[Hochreiter et~al.(2001)Hochreiter, Bengio, Frasconi, Schmidhuber,
  et~al.]{hochreiter2001gradient}
Sepp Hochreiter, Yoshua Bengio, Paolo Frasconi, J{\"u}rgen Schmidhuber, et~al.
\newblock Gradient flow in recurrent nets: the difficulty of learning long-term
  dependencies, 2001.

\bibitem[Hwang et~al.(2015)Hwang, Lee, and Sung]{HWang}
Kyuyeon Hwang, Minjae Lee, and Wonyong Sung.
\newblock Online keyword spotting with a character-level recurrent neural
  network.
\newblock \emph{arXiv:1512.08903}, 2015.

\bibitem[Itakura(1990)]{Fumitada90dtw}
Fumitada Itakura.
\newblock \emph{Minimum Prediction Residual Principle Applied to Speech
  Recognition}, page 154–158.
\newblock Morgan Kaufmann Publishers Inc., San Francisco, CA, USA, 1990.
\newblock ISBN 1558601244.

\bibitem[Jha et~al.(2018)Jha, Namboodiri, and Jawahar]{Jha}
Abhishek Jha, Vinay~P. Namboodiri, and C.~V. Jawahar.
\newblock Word spotting in silent lip videos.
\newblock In \emph{IEEE Winter Conference on Applications of Computer Vision},
  2018.

\bibitem[K~R et~al.(2021)K~R, Afouras, Zisserman, et~al.]{prajwal2021sub}
Prajwal K~R, Triantafyllos Afouras, Andrew Zisserman, et~al.
\newblock Sub-word level lip reading with visual attention.
\newblock \emph{arXiv preprint arXiv:2110.07603}, 2021.

\bibitem[Karita et~al.(2019)Karita, Yalta, Watanabe, Delcroix, Ogawa, and
  Nakatani]{Karita2019ImprovingTE}
Shigeki Karita, Nelson Yalta, Shinji Watanabe, M.~Delcroix, A.~Ogawa, and
  T.~Nakatani.
\newblock Improving transformer-based end-to-end speech recognition with
  connectionist temporal classification and language model integration.
\newblock In \emph{INTERSPEECH}, 2019.

\bibitem[Kim and Nam(2019)]{Kim}
Taejun Kim and Juhan Nam.
\newblock Temporal feedback convolutional recurrent neural networks for keyword
  spotting.
\newblock \emph{arXiv:1911.01803}, 2019.

\bibitem[Kingma and Ba(2015)]{kingma2014adam}
Diederik~P Kingma and Jimmy Ba.
\newblock {ADAM}: A method for stochastic optimization.
\newblock In \emph{Proc. ICLR}, 2015.

\bibitem[Lengerich and Hannun(2016)]{Lengerich}
Chris Lengerich and Awni Hannun.
\newblock An end-to-end architecture for keyword spotting and voice activity
  detection.
\newblock In \emph{NIPS 2016 End-to-End Learning for Speech and Audio
  Processing Workshop}, 2016.

\bibitem[Li et~al.(2019)Li, Wang, Lei, Siniscalchi, and Lee]{li2019improving}
Wei Li, Sicheng Wang, Ming Lei, Sabato~Marco Siniscalchi, and Chin-Hui Lee.
\newblock Improving audio-visual speech recognition performance with
  cross-modal student-teacher training.
\newblock In \emph{Proc. ICASSP}, pages 6560--6564. IEEE, 2019.

\bibitem[Liu et~al.(2018)Liu, Wang, Nie, He, Chen, and
  Chua]{Liu2018AttentiveMR}
Meng Liu, Xiang Wang, Liqiang Nie, Xiangnan He, Baoquan Chen, and Tat-Seng
  Chua.
\newblock Attentive moment retrieval in videos.
\newblock In \emph{The 41st International ACM SIGIR Conference on Research \&
  Development in Information Retrieval}, 2018.

\bibitem[Makino et~al.(2019)Makino, Liao, Assael, Shillingford, Garcia, Braga,
  and Siohan]{makino2019recurrent}
Takaki Makino, Hank Liao, Yannis Assael, Brendan Shillingford, Basilio Garcia,
  Otavio Braga, and Olivier Siohan.
\newblock Recurrent neural network transducer for audio-visual speech
  recognition.
\newblock In \emph{IEEE Workshop on Automatic Speech Recognition and
  Understanding}, 2019.

\bibitem[Mohamed et~al.(2019)Mohamed, Okhonko, and
  Zettlemoyer]{mohamed2019transformers}
Abdelrahman Mohamed, Dmytro Okhonko, and Luke Zettlemoyer.
\newblock Transformers with convolutional context for asr.
\newblock \emph{arXiv preprint arXiv:1904.11660}, 2019.

\bibitem[Momeni et~al.(2020{\natexlab{a}})Momeni, Afouras, Stafylakis, Albanie,
  and Zisserman]{Momeni20}
Liliane Momeni, Triantafyllos Afouras, Themos Stafylakis, Samuel Albanie, and
  Andrew Zisserman.
\newblock Seeing wake words: Audio-visual keyword spotting.
\newblock In \emph{BMVC}, 2020{\natexlab{a}}.

\bibitem[Momeni et~al.(2020{\natexlab{b}})Momeni, Varol, Albanie, Afouras, and
  Zisserman]{Momeni20b}
Liliane Momeni, G{\"u}l Varol, Samuel Albanie, Triantafyllos Afouras, and
  Andrew Zisserman.
\newblock Watch, read and lookup: Learning to spot signs from multiple
  supervisors.
\newblock In \emph{Proc. ACCV}, 2020{\natexlab{b}}.

\bibitem[Palaz et~al.(2016)Palaz, Synnaeve, and Collobert]{Palaz}
Dimitri Palaz, Gabriel Synnaeve, and Ronan Collobert.
\newblock Jointly learning to locate and classify words using convolutional
  networks.
\newblock In \emph{INTERSPEECH}, 2016.

\bibitem[Papandreou et~al.(2009)Papandreou, Katsamanis, Pitsikalis, and
  Maragos]{papandreou09}
George Papandreou, Athanassios Katsamanis, Vassilis Pitsikalis, and Petros
  Maragos.
\newblock Adaptive multimodal fusion by uncertainty compensation with
  application to audiovisual speech recognition.
\newblock \emph{Audio, Speech, and Language Processing, IEEE Transactions on},
  17\penalty0 (3):\penalty0 423--435, 2009.

\bibitem[Petridis et~al.(2018)Petridis, Stafylakis, Ma, Tzimiropoulos, and
  Pantic]{petridis2018audio}
Stavros Petridis, Themos Stafylakis, Pingchuan Ma, Georgios Tzimiropoulos, and
  Maja Pantic.
\newblock {Audio-Visual Speech Recognition with a Hybrid CTC/Attention
  Architecture}.
\newblock In \emph{IEEE Spoken Language Technology Workshop}, pages 513--520,
  2018.

\bibitem[Potamianos et~al.(2003)Potamianos, Neti, Gravier, Garg, and
  Senior]{potamianos2003recent}
Gerasimos Potamianos, Chalapathy Neti, Guillaume Gravier, Ashutosh Garg, and
  Andrew~W Senior.
\newblock Recent advances in the automatic recognition of audiovisual speech.
\newblock \emph{Proceedings of the IEEE}, 2003.

\bibitem[Radford et~al.(2019)Radford, Wu, Child, Luan, Amodei, and
  Sutskever]{radford2019language}
Alec Radford, Jeffrey Wu, Rewon Child, David Luan, Dario Amodei, and Ilya
  Sutskever.
\newblock Language models are unsupervised multitask learners.
\newblock \emph{OpenAI blog}, 1\penalty0 (8):\penalty0 9, 2019.

\bibitem[Rose and Paul(1990)]{rose1990hidden}
Richard~C Rose and Douglas~B Paul.
\newblock A hidden markov model based keyword recognition system.
\newblock In \emph{International Conference on Acoustics, Speech, and Signal
  Processing}, pages 129--132. IEEE, 1990.

\bibitem[Rosenberg et~al.(2017)Rosenberg, Audhkhasi, Sethy, Ramabhadran, and
  Picheny]{rosenberg2017end}
Andrew Rosenberg, Kartik Audhkhasi, Abhinav Sethy, Bhuvana Ramabhadran, and
  Michael Picheny.
\newblock End-to-end speech recognition and keyword search on low-resource
  languages.
\newblock In \emph{ICASSP}, 2017.

\bibitem[Sainath and Parada(2015)]{Sainath2015ConvolutionalNN}
Tara~N. Sainath and Carolina Parada.
\newblock Convolutional neural networks for small-footprint keyword spotting.
\newblock In \emph{INTERSPEECH}, 2015.

\bibitem[Sakoe and Chiba(1978)]{Sakoe78dynamicprogramming}
Hiroaki Sakoe and Seibi Chiba.
\newblock Dynamic programming algorithm optimization for spoken word
  recognition.
\newblock \emph{IEEE TRANSACTIONS ON ACOUSTICS, SPEECH, AND SIGNAL PROCESSING},
  26:\penalty0 43--49, 1978.

\bibitem[Schembri et~al.(2013)Schembri, Fenlon, Rentelis, Reynolds, and
  Cormier]{schembri2013building}
Adam Schembri, Jordan Fenlon, Ramas Rentelis, Sally Reynolds, and Kearsy
  Cormier.
\newblock Building the {B}ritish {S}ign {L}anguage {C}orpus.
\newblock \emph{Language Documentation \& Conservation}, 7:\penalty0 136--154,
  2013.

\bibitem[Schembri et~al.(2017)Schembri, Fenlon, Rentelis, and
  Cormier]{bslcorpus17}
Adam Schembri, Jordan Fenlon, Ramas Rentelis, and Kearsy Cormier.
\newblock {British Sign Language Corpus Project: A corpus of digital video data
  and annotations of British Sign Language 2008-2017 (Third Edition)}, 2017.
\newblock URL \url{http://www.bslcorpusproject.org}.

\bibitem[Shillingford et~al.(2019)Shillingford, Assael, Hoffman, Paine, Hughes,
  Prabhu, Liao, Sak, Rao, Bennett, Mulville, Coppin, Laurie, Senior, and
  de~Freitas]{Shillingford18}
Brendan Shillingford, Yannis Assael, Matthew~W. Hoffman, Thomas Paine, Cían
  Hughes, Utsav Prabhu, Hank Liao, Hasim Sak, Kanishka Rao, Lorrayne Bennett,
  Marie Mulville, Ben Coppin, Ben Laurie, Andrew Senior, and Nando de~Freitas.
\newblock {Large-Scale Visual Speech Recognition}.
\newblock In \emph{INTERSPEECH}, 2019.

\bibitem[{Speech Group at Carnegie Mellon University}(2014)]{cmu}
{Speech Group at Carnegie Mellon University}.
\newblock {{CMU}} pronouncing dictionary.
\newblock \url{http://www.speech.cs.cmu.edu/cgi-bin/cmudict}, 2014.

\bibitem[Stafylakis and Tzimiropoulos(2017)]{Stafylakis17}
Themos Stafylakis and Georgios Tzimiropoulos.
\newblock Combining residual networks with lstms for lipreading.
\newblock In \emph{INTERSPEECH}, 2017.

\bibitem[Stafylakis and Tzimiropoulos(2018)]{stafylakis18kws}
Themos Stafylakis and Georgios Tzimiropoulos.
\newblock Zero-shot keyword spotting for visual speech recognition in-the-wild.
\newblock In \emph{ECCV}, 2018.

\bibitem[Sun et~al.(2016)Sun, Raju, Tucker, Panchapagesan, Fu, Mandal,
  Matsoukas, Strom, and Vitaladevuni]{Sun2016MaxpoolingLT}
Ming Sun, Anirudh Raju, George Tucker, Sankaran Panchapagesan, Gengshen Fu,
  Arindam Mandal, Spyridon Matsoukas, Nikko Strom, and Shiv Vitaladevuni.
\newblock Max-pooling loss training of long short-term memory networks for
  small-footprint keyword spotting.
\newblock \emph{2016 IEEE Spoken Language Technology Workshop (SLT)}, 2016.

\bibitem[Sutton-Spence(2007)]{sutton2007mouthings}
Rachel Sutton-Spence.
\newblock Mouthings and simultaneity in british sign language.
\newblock \emph{Amsterdam studies in the theory and history of linguistic
  science, series 4}, 281:\penalty0 147, 2007.

\bibitem[Varol et~al.(2021)Varol, Momeni, Albanie, Afouras, and
  Zisserman]{varol21}
G{\"u}l Varol, Liliane Momeni, Samuel Albanie, Triantafyllos Afouras, and
  Andrew Zisserman.
\newblock Read and attend: Temporal localisation in sign language videos.
\newblock In \emph{CVPR}, 2021.

\bibitem[Vaswani et~al.(2017)Vaswani, Shazeer, Parmar, Uszkoreit, Jones, Gomez,
  Kaiser, and Polosukhin]{vaswani2017attention}
Ashish Vaswani, Noam Shazeer, Niki Parmar, Jakob Uszkoreit, Llion Jones,
  Aidan~N Gomez, {\L}ukasz Kaiser, and Illia Polosukhin.
\newblock Attention is all you need.
\newblock In \emph{NeurIPS}, 2017.

\bibitem[Wang et~al.(2017)Wang, Getreuer, Hughes, Lyon, and Saurous]{Wang2}
Yuxuan Wang, Pascal Getreuer, Thad Hughes, Richard Lyon, and Rif Saurous.
\newblock Trainable frontend for robust and far-field keyword spotting.
\newblock In \emph{ICASSP}, 2017.

\bibitem[Wilpon et~al.(1989)Wilpon, Lee, and Rabiner]{Wilpon89}
Jay Wilpon, Chin-Hui Lee, and Lawrence Rabiner.
\newblock Application of hidden markov models for recognition of a limited set
  of words in unconstrained speech.
\newblock In \emph{International Conference on Acoustics, Speech, and Signal
  Processing,}, pages 254 -- 257 vol.1, 06 1989.
\newblock \doi{10.1109/ICASSP.1989.266413}.

\bibitem[Wu et~al.(2020)Wu, Xu, Dai, Wan, Zhang, Tomizuka, Keutzer, and
  Vajda]{wu20visual}
Bichen Wu, Chenfeng Xu, Xiaoliang Dai, Alvin Wan, Peizhao Zhang, Masayoshi
  Tomizuka, Kurt Keutzer, and Peter Vajda.
\newblock Visual transformers: Token-based image representation and processing
  for computer vision.
\newblock \emph{CoRR}, abs/2006.03677, 2020.

\bibitem[Wu et~al.(2016)Wu, Liu, Li, Fan, and Zhang]{Wu16avkws}
Pingping Wu, Hong Liu, Xiaofei Li, Ting Fan, and Xuewu Zhang.
\newblock A novel lip descriptor for audio-visual keyword spotting based on
  adaptive decision fusion.
\newblock \emph{IEEE Transactions on Multimedia}, 2016.

\bibitem[Xu et~al.(2019)Xu, He, Plummer, Sigal, Sclaroff, and
  Saenko]{Xu2019MultilevelLA}
Huijuan Xu, Kun He, Bryan~A. Plummer, L.~Sigal, S.~Sclaroff, and Kate Saenko.
\newblock Multilevel language and vision integration for text-to-clip
  retrieval.
\newblock In \emph{AAAI}, 2019.

\bibitem[Yao et~al.(2019)Yao, Wang, Du, Zheng, and Gedeon]{sliding-windows}
Yue Yao, Tianyu Wang, Heming Du, Liang Zheng, and Tom Gedeon.
\newblock Spotting visual keywords from temporal sliding windows.
\newblock In \emph{Mandarin Audio-Visual Speech Recognition Challenge}, 2019.

\bibitem[Yu et~al.(2020)Yu, Zhang, Wu, Ghorbani, Wu, Kang, Liu, Liu, Meng, and
  Yu]{Jianwei20tdnn}
Jianwei Yu, Shi-Xiong Zhang, Jian Wu, Shahram Ghorbani, Bo~Wu, Shiyin Kang,
  Shansong Liu, Xunying Liu, Helen Meng, and Dong Yu.
\newblock Audio-visual recognition of overlapped speech for the lrs2 dataset.
\newblock In \emph{ICASSP}, pages 6984--6988, 05 2020.
\newblock \doi{10.1109/ICASSP40776.2020.9054127}.

\bibitem[Yuan et~al.(2019)Yuan, Mei, and Zhu]{Yuan2019ToFW}
Yitian Yuan, T.~Mei, and Wenwu Zhu.
\newblock To find where you talk: Temporal sentence localization in video with
  attention based location regression.
\newblock In \emph{AAAI}, 2019.

\bibitem[Zeng et~al.(2020)Zeng, Xu, Huang, Chen, Tan, and Gan]{Zeng2020DenseRN}
Runhao Zeng, H.~Xu, W.~Huang, Peihao Chen, Mingkui Tan, and Chuang Gan.
\newblock Dense regression network for video grounding.
\newblock In \emph{CVPR}, 2020.

\bibitem[Zhang et~al.(2018)Zhang, Zhang, and Wang]{Zhang18}
Haitong Zhang, Junbo Zhang, and Yujun Wang.
\newblock Sequence-to-sequence models for small-footprint keyword spotting.
\newblock \emph{arXiv:1811.00348}, 2018.

\bibitem[Zhang et~al.(2019)Zhang, Cheng, and Wang]{zhang2019spatio}
Xingxuan Zhang, Feng Cheng, and Shilin Wang.
\newblock Spatio-temporal fusion based convolutional sequence learning for lip
  reading.
\newblock In \emph{Proc. ICCV}, pages 713--722, 2019.

\bibitem[Zhang and Glass(2010)]{Yaodong10}
Yaodong Zhang and James Glass.
\newblock Unsupervised spoken keyword spotting via segmental dtw on gaussian
  posteriorgrams.
\newblock In \emph{Proceedings of the 2009 IEEE Workshop on Automatic Speech
  Recognition and Understanding}, pages 398 -- 403, 01 2010.
\newblock \doi{10.1109/ASRU.2009.5372931}.

\bibitem[Zhang et~al.(2017)Zhang, Suda, Lai, and Chandra]{HelloEdge}
Yundong Zhang, Naveen Suda, Liangzhen Lai, and Vikas Chandra.
\newblock Hello edge: Keyword spotting on microcontrollers.
\newblock \emph{CoRR}, 2017.

\bibitem[Zhou et~al.(2014)Zhou, Zhao, Hong, and Pietik{\"a}inen]{Zhou14}
Ziheng Zhou, Guoying Zhao, Xiaopeng Hong, and Matti Pietik{\"a}inen.
\newblock A review of recent advances in visual speech decoding.
\newblock \emph{Image and vision computing}, 32\penalty0 (9):\penalty0
  590--605, 2014.

\bibitem[Zhuang et~al.(2016)Zhuang, Chang, Qian, and
  Yu]{Zhuang2016UnrestrictedVK}
Yimeng Zhuang, Xuankai Chang, Yanmin Qian, and Kai Yu.
\newblock {Unrestricted Vocabulary Keyword Spotting Using LSTM-CTC}.
\newblock In \emph{INTERSPEECH}, 2016.

\end{thebibliography}
\clearpage
\title{APPENDIX}
\author{}
\maketitle

\appendix
\section{Application to silent films}
\label{supp:sec:silent}
Our Transpotter model has been trained to spot words in silent talking face videos. Naturally, the next step is to apply it on scenes from actual silent films to detect if a word is spoken and if so, also localize it in time. On the project website, we show qualitative examples of applying the Transpotter on scenes from the silent film, \href{https://en.wikipedia.org/wiki/The_Artist_(film)}{The Artist}.
Silent films often use title cards to convey character dialogues. Although audio was not recorded, actors would still talk during the shooting, following the script, as it enabled them to act more naturally. We therefore used the film's screenplay (available online) to identify scenes where title cards appear, which indicates the presence of dialogue, and queried words out of the title card content of these scenes. This application is particularly challenging because
the title cards are only a guideline for the dialogue, therefore the lines that the actors end up using may differ substantially. Moreover we note that in this particular example, some of the actors are not native English speakers, therefore the model also has to deal with a domain shift in terms of accent.
Despite those challenges, the Transpotter is still able to detect and properly localize words. Another application with a similar domain shift is spotting mouthings in sign language, which we have already discussed in the main paper.
\vspace{10pt}

The rest of this PDF is organized as follows. Section~\ref{supp:sec:datasets} describes the datasets we used for training and evaluation in detail. Next, we elaborate on the training process for the VSR baseline in Section~\ref{supp:sec:lipreading}. In Section~\ref{supp:sec:hyperparams}, we describe the hyper-parameters choices for the Transpotter architecture.
In Section~\ref{supp:sec:phrases}, we evaluate the model for longer query sequences containing multiple words.
In Section~\ref{supp:sec:arch_ablations}, we provide details for the architecture variants that were used in the main paper. In Section~\ref{supp:sec:ablations}, we conduct two more ablation studies, on the choice of the localization head and on the need for modality embeddings. 

\section{Dataset Statistics}
\label{supp:sec:datasets}

We conduct most of our experiments on the Lip Reading LRS2~\cite{Chung16b,Afouras19} and LRS3~\cite{Afouras18d} datasets. Each dataset contains three splits: (i) Pre-train, (ii) Train-val, and (iii) Test. The utterances in the pre-train set correspond to part-sentences as well as multiple sentences, whereas the training set only consists of single full sentences or phrases. The number of utterances, words and vocabulary for each of the splits in both datasets is reported in Table~\ref{tab:datasets}.

\begin{table}[ht]
    \setlength{\tabcolsep}{10pt}
    \centering
    \resizebox{0.999\linewidth}{!}{
        \begin{tabular}{c c c c c c }
            \toprule
             Dataset & Split & \#Utterances & \#Words & \#Hours  & Vocabulary \\ 
            \midrule
            & Pre-train & 96k & 2M &195 & 41k \\
            LRS2~\cite{Chung16b, Afouras19} & Train-val & 47k & 336k &29 & 18k \\
             & Test      & 1.2k & 6k &0.5 & 1.7k \\
             \midrule
             & Pre-train & 132k & 3.9M	& 444 & 51k  \\
            LRS3~\cite{Afouras18d} & Train-val & 32k & 358k&30 & 17k\\
            & Test      & 1.3k & 10k &1 &2k \\
            \bottomrule
        \end{tabular}
    }
    \vspace{2pt}
     \caption{\textbf{Audio-visual datasets}: Statistics of LRS2 and LRS3 datasets.}
    \label{tab:datasets}
\end{table}

Our models are first pre-trained on both the LRS2, LRS3 pre-train splits. Before testing on the test set of each dataset, we fine-tune the model on the train-val split of that particular dataset. Training hyper-parameters are detailed in the next sections.

\section{VSR baseline}
\label{supp:sec:lipreading}

In this section, we describe the exact architectural details and training hyper-parameters for the VSR baseline. 

\noindent
\textbf{Architecture details.}
The architecture closely resembles the TM-seq2seq architecture~\cite{Afouras19}. The only minor changes are all in the visual backbone, which we describe below in Table~\ref{app:tab:cnn_arch}. 

\begin{table}[ht]
  \centering

  \vspace{-5pt}
  \hspace{-12pt}
  \begin{tabular}[t]{  l r c c c r }
   \toprule
   Layer & \# channels & Kernel & Stride & Padding  & Output dimensions  \\  
   \midrule
   input  & 3 &   - &  -      &   -    & $T \times 112 \times 112$  \\  
   conv\textsubscript{1,1}  & 64 & (5,5,5) & (1,2,2) & (2,2,2)  & $T \times 56 \times 56 $  \\   
   \midrule
   conv\textsubscript{2,1} & 128 & (3,3)   & (2,2) &  (1,1)  &   $T \times 28 \times 28 $  \\ 
   conv\textsubscript{2,2} & 128 & (3,3)   & (1,1) &  (1,1)  &   $T \times 28 \times 28 $  \\ 
   conv\textsubscript{2,3} & 128 & (3,3)   & (1,1) &  (1,1)  &   $T \times 28 \times 28 $  \\ 
   \midrule
   conv\textsubscript{3,1} & 256 & (3,3)   & (2,2) &  (1,1)  &   $T \times 14 \times 14 $  \\ 
   conv\textsubscript{3,2} & 256 & (3,3)   & (1,1) &  (1,1)  &   $T \times 14 \times 14 $  \\ 
   conv\textsubscript{3,3} & 256 & (3,3)   & (1,1) &  (1,1)  &   $T \times 14 \times 14 $  \\ 
   \midrule
   conv\textsubscript{4,1} & 512 & (3,3)   & (2,2) &  (1,1)  &   $T \times 7 \times 7 $  \\ 
   conv\textsubscript{4,2} & 512 & (3,3)   & (1,1) &  (1,1)  &   $T \times 7 \times 7 $  \\ 
   conv\textsubscript{4,3} & 512 & (3,3)   & (1,1) &  (1,1)  &   $T \times 7 \times 7 $  \\ 
  \midrule
   conv\textsubscript{5,1} & 512 & (3,3)   & (2,2) &  (1,1)  &  $T \times 4 \times 4 $  \\ 
   fc                      & 512 & (4,4)   & (1,1) &  (0,0)  &   $T \times 1 \times 1 $  \\ 
   \bottomrule
  \end{tabular}

  \normalsize
    \vspace{5pt}
  \caption{
    \textbf{Architecture details for the visual CNN backbone.}
    Batch Normalization and ReLU activation are added after every convolutional layer. 
    Shortcut connections are also added at each layer, except for the first layer of every residual block -- i.e. the ones with stride 2 (conv\textsubscript{1,1}, conv\textsubscript{2,1} and conv\textsubscript{3,1}, conv\textsubscript{4,1}).
  }
  \label{app:tab:cnn_arch}

\end{table}

\noindent
\textbf{Training hyper-parameters.} We train the VSR baseline on LRS2, LRS3. The training consists of two stages, similar to~\cite{Afouras19}. 

In the first stage, we train the visual backbone end-to-end with the transformer layers on all two-word video clips (obtained using the available LRS2, LRS3 word alignments). Both the CNN and Transformer layers use the Adam optimizer~\cite{kingma2014adam}, but with different learning rate schedules. For the CNN, we  start with an initial learning rate of $1e^{-4}$ and reduce it by a factor of $2$ every time the validation loss plateaus for $3$ epochs. The minimum learning rate for the CNN is $1e^{-5}$. The transformer layers start with an initial learning rate of $5e^{-5}$. We do not reduce this learning rate in the first stage. The first stage takes approximately $6$ days on four Tesla v100 GPUs.

After the visual backbone is trained, we extract the features for all the video clips in our datasets, and use these fixed features for all further training. We follow the curriculum strategy of~\cite{Afouras19} and train the transformer layers for longer video segments ($9$ words). We halve the learning rate every time the validation loss plateaus for $10$ epochs. The minimum learning rate for the transformer layers is $1e^{-6}$. The second stage takes about $2$ days on one Tesla v100 GPU. 

Similar to the spotting models, we fine-tune on the train-val set of LRS2/LRS3 before evaluation. During inference, we decode text sequences with a left-to-right beam search with a beam width of $B=20$. The beam hypotheses scores and textual outputs are used for computing the keyword spotting scores. More details of the evaluation mechanism can be found in this paper~\cite{he}.

\section{Transpotter hyperparameters}
\label{supp:sec:hyperparams}

\textbf{Model hyper-parameters.} We set the number of text encoder layers $N_t = 3$, the number of video encoder layers $N_v = 6$, and the number of joint multimodal transformer layers $T_m = 6$. The embedding dimension $d$ is set to $512$, and each multi-headed attention layer uses $8$ heads.

\noindent
\textbf{Training hyper-parameters.} The Transpotter is trained on the extracted visual features that we obtain after Stage 1 training (described above) of the VSR model. We use the Adam optimizer~\cite{kingma2014adam} with an initial learning rate of $5e^{-5}$, which is decreased by a factor of $5$ every time the validation loss plateaus for $15$ epochs. The minimum learning rate is $1e^{-6}$. The model is trained for $280$ epochs on a single Ampere A40 GPU, which takes about $4 - 5$ days. We fine-tune on the train-val set of LRS2/LRS3 before evaluation.

\section{Using the Transpotter to spot phrases} \label{supp:sec:phrases}

The Transpotter architecture can be used to spot multiple words (phrases) without any change in the architecture. Instead of feeding a phoneme sequence for a single word, we simply feed a phoneme sequence for a phrase. This is obtained by concatenating the phoneme sequence of each word in the phrase. We fine-tune the word-level model for a slight boost in performance. 

In order to evaluate this model, we form our query set by using n-grams from the LRS2 and LRS3 test sets. We evaluate for $n = {1,2,3}$. Note that $n = 1$ corresponds to single words. As is done for our single word evaluation in the main paper, we only use words with number of phonemes $n_p \geq 3$. The scores are reported in Table~\ref{supp:tab:phrases}. We know that Transpotter becomes increasingly accurate with longer query lengths from Figure 2 (a) of the main paper, and the same trend continues when we try to spot phrases instead of single words. Spotting phrases is clearly far easier than spotting single words. 

\begin{table}[ht]
    \setlength{\tabcolsep}{8pt}
    \centering
    \resizebox{0.99\linewidth}{!}{
        \begin{tabular}{l|ccc|c|ccc|c}
            \toprule
             & \multicolumn{4}{c}{LRS2} & \multicolumn{4}{|c}{LRS3}\\
            \midrule
            Num. words  & \aconecls & \acfivecls  & \mapcls  &\maploc & \aconecls & \acfivecls  & \mapcls   &\maploc \\
            \midrule
            n = 1  &  65.0 & 87.1 &  69.2 & 68.3  & 52.0 & 77.1 & 55.4  & 53.6 \\
             n = 2  &  74.9 & 94.6 &  82.9 & 82.7  & 57.7 & 82.8 & 67.8 &  66.9\\
            n = 3  &  87.9 & 99.4 &  92.9 & 92.7  & 74.7 & 91.0 & 82.0  &  80.6 \\
            \bottomrule
        \end{tabular}
    }
    \vspace{2pt}
    \caption{\textbf{Spotting phrases:} Transpotter can be trained to spot multi-word queries (phrases) with no change in the architecture. We report the scores for spotting unigrams (single words), bigrams and trigrams. Longer queries can be spotted far more easily.
    }
    \label{supp:tab:phrases}
    \ptab
\end{table}

\section{Details for the Architecture Ablations}
\label{supp:sec:arch_ablations}

\begin{figure}[ht]
    \centering
    \includegraphics[width=\textwidth]{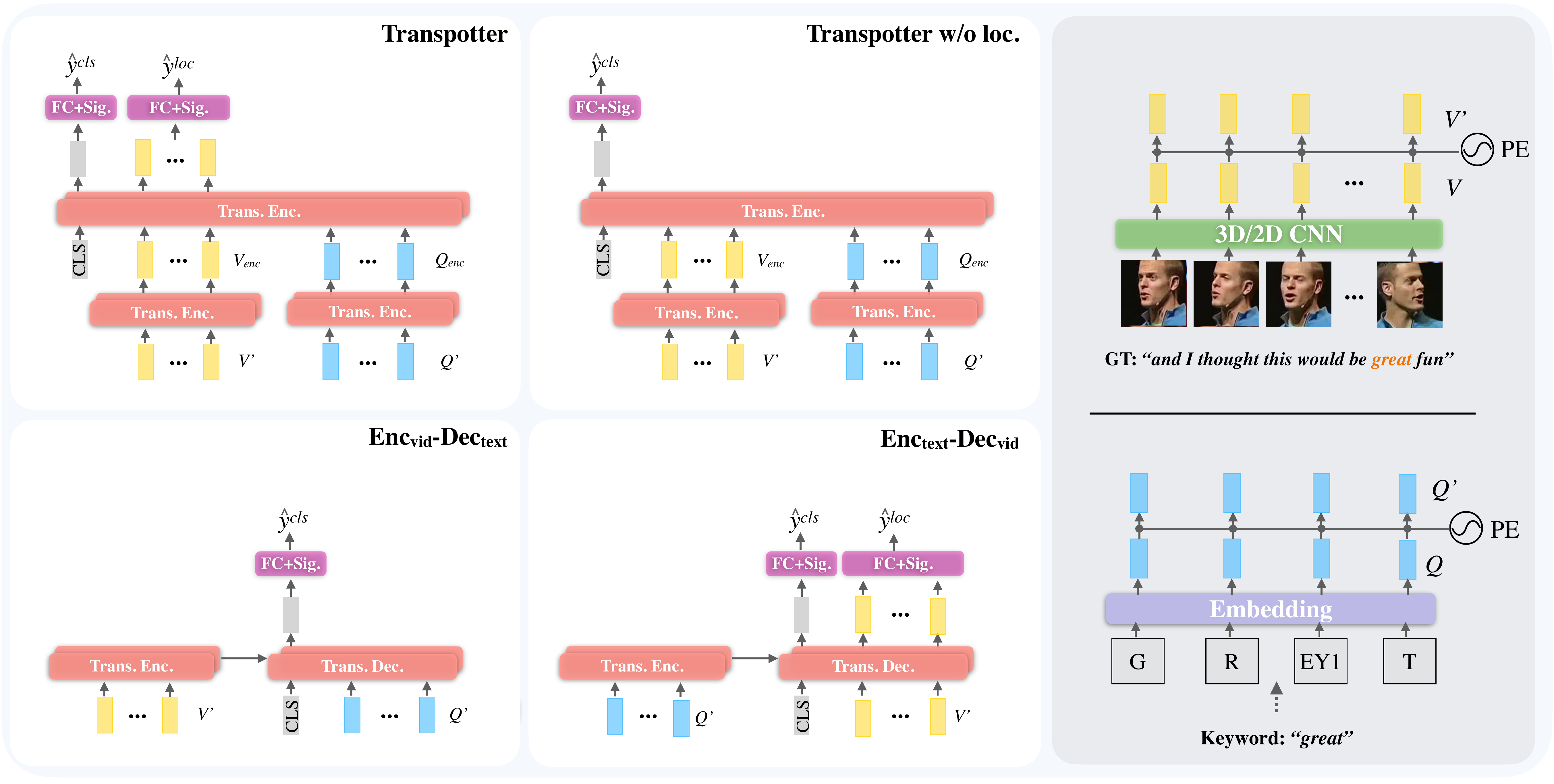}
    \caption{\textbf{Architecture ablations:} We illustrate each of the architecture variants tried for this task. The models are compared in Table 2 of the main paper.
    }
    \label{fig:model_ablations}
\end{figure}

In Table 2 (Section 4.3) of the main paper, we experiment with alternative Transformer architectures for this task. We explain each of the architectures below and also illustrate them in Figure~\ref{fig:model_ablations}. 

\noindent
\textbf{$\text{Enc}_{vid}$-$\text{Dec}_{text}$:} This is the standard Transformer Encoder-Decoder architecture, which has also been used with great success~\cite{Afouras19} in related tasks such as VSR. The video encoder contains $6$ Transformer layers encoding the temporal information for the video input. A text decoder consists of $6$ Transformer decoder layers, each with (i) self-attention across the phoneme feature vectors, (ii) cross-attention between the phoneme and video feature vectors. The text input to the decoder is preprended with a [CLS] vector, where the probability of the keyword occurring  in the video is predicted. Note that this model cannot explicitly localize the keyword in the video, as the output time-steps correspond to the text sequence and not the video.

\noindent
\textbf{$\text{Enc}_{text}$-$\text{Dec}_{vid}$:} This architecture is similar to the previous one, except the inputs are swapped. The encoder processes the text sequence and the decoder inputs the video frames (prepended with a [CLS] vector). Since the output time-steps of the decoder correspond to the video, the exact same localization loss used in Transpotter can be applied here for localizing the keyword. 

\noindent
\textbf{Transpotter w/o localization head:} 
We also train our Transpotter model without the localization head. We only optimize for the presence/absence of the keyword at the [CLS] time-step. 

All the above variants are trained using the exact same data pipeline and optimizer hyper-parameters.

\section{Additional Ablations}
\label{supp:sec:ablations}
\subsection{Choice of Localization Head}

In Table~\ref{supp:tab:loss_ablations}, we compare our proposed localization head with an alternative start-end span prediction head. For this, we have two MLP layers that predict start and end probabilities at each video frame indicating the span of the keyword. The frame-wise probabilities are softmax-normalized and optimized with cross-entropy loss during training. We observe that our proposed localization head consistently outperforms this alternative technique.

\begin{table}[ht]
    \setlength{\tabcolsep}{8pt}
    \centering
    \resizebox{0.999\linewidth}{!}{
        \begin{tabular}{l|ccc|c|ccc|c}
            \toprule
             & \multicolumn{4}{c}{LRS2} & \multicolumn{4}{|c}{LRS3}\\
            \midrule
              Loss & \aconecls & \acfivecls  & \mapcls &\maploc & \aconecls & \acfivecls  & \mapcls  &\maploc \\
            \midrule
            \losscls + \losslocsoftmax &  61.9 & 85.5 & 66.6 & 65.9 & 49.9 & 74.8  &  52.8 & 51.6 \\
            \losscls + \losslocbce  & \textbf{ 65.0} & \textbf{87.1} & \textbf{69.2 }& \textbf{68.3 } & \textbf{ 52.0 }& \textbf{77.1 }&\textbf{ 55.4 } &\textbf{ 53.6} \\
            \bottomrule
        \end{tabular}
    }
     \label{supp:tab:loss_ablations}
    \caption{\textbf{Ablations on the Localization head:} Per-frame sigmoid prediction of the keyword location outperforms the softmax-based span prediction.
}
\end{table}

\subsection{Are modality embeddings needed?}
\begin{table}[ht]
    \setlength{\tabcolsep}{8pt}
    \centering
    \resizebox{0.999\linewidth}{!}{
        \begin{tabular}{c|ccc|c|ccc|c}
            \toprule
             & \multicolumn{4}{c}{LRS2} & \multicolumn{4}{|c}{LRS3}\\
            \midrule
              Modality Tokens & \aconecls & \acfivecls  & \mapcls &\maploc & \aconecls & \acfivecls  & \mapcls  &\maploc \\
            \midrule
            $\checkmark$ &  64.6 & \textbf{87.3} & 68.9 & 68.3 & \textbf{52.9} & 77.0  &  55.4 & 53.5 \\
            $\times$  & \textbf{ 65.0} & 87.1 & \textbf{69.2 }& \textbf{68.3 } &  52.0 & \textbf{77.1 }&\textbf{ 55.4 } &\textbf{ 53.6} \\
            \bottomrule
        \end{tabular}
    }
     \label{supp:tab:mod_tokens}
    \caption{\textbf{Are modality embeddings needed in Transpotter?} Presence/absence of modality embeddings to explicitly specify the text and video time-steps does not affect the performance. Our dedicated video and text encoders encode sufficient modality specific information.
}
\end{table}

Since we concatenate our video and text embeddings together into a single sequence for feeding into the joint multi-modal transformer, the question arises if we need to explicitly specify which time-steps are video/text. We argue that this is not necessary because we have separate text and video encoders before the joint transformer. In order to empirically show this, we train a model with learnable modality tokens that are added to the text and video time-steps. We use two learnable vectors (one each for video/text) of embedding dimension $d$ and add them to each of the video/text time-step accordingly. We perform this addition just before feeding them into the joint transformer. In Table~\ref{supp:tab:mod_tokens}, we show that this performs comparably to our Transpotter model without any modality tokens.

\section{Error analysis on LRW}
In this section, we illustrate some of the common errors made by our model. We conduct this analysis on the LRW test set using the same retrieval protocol that we have used in all our experiments, i.e.~given a query word, we retrieve the top-K videos. Since LRW videos correspond to a single word label, we can clearly see which word labels are wrongly retrieved. 

In Table~\ref{supp:tab:lrw_analysis}, we show the word labels of the incorrectly retrieved videos (among the top 10 retrievals) for some of the query words. We observe that often errors occur due to some of the phonemes of the query word appearing in the video. For example, the words ``president'' and ``prison'' share several phonemes. We also observe that compound words are a source of error: for example, in the second row, where ``every'' appears entirely in ``everybody'' and ``everything''. Finally, as discussed in Section~\ref{subsec:analysis}, another failure case are homophemes. For example,  in the last row, we see an example of the model retrieving the homopheme ``million" for the query word ``billion''. 

\begin{table}[ht]
    \setlength{\tabcolsep}{8pt}
    \centering
        \begin{tabular}{l|r}
            \toprule
            Query word & Mistakes among the top $10$ retrievals\\
            \midrule
            president & press, prison, pressure \\
            every  & everybody, everything \\
            allowed & allow, cloud, announced \\
            example & couple, happen \\
            billion & million, building \\
            \bottomrule
        \end{tabular}
    \vspace{3mm}
     \label{supp:tab:lrw_analysis}
    \caption{\textbf{Error analysis on LRW:} For each query word, we report examples of incorrectly retrieved word instances among the top 10 retrievals. We can see clear patterns in mistakes such as phoneme overlap between the query word and the mistaken word, compound words and homophemes. 
}
\end{table}

\end{document}